# Kernel-Induced Label Propagation by Mapping for Semi-Supervised Classification

Zhao Zhang, *Senior Member*, *IEEE*, Lei Jia, Mingbo Zhao, *Member*, *IEEE*, Guangcan Liu, *Member*, *IEEE*, Meng Wang, *Member*, *IEEE*, and Shuicheng Yan, *Fellow*, *IEEE*

**Abstract**— Kernel methods have been successfully applied to the areas of pattern recognition and data mining. In this paper, we mainly discuss the issue of propagating labels in kernel space. A Kernel-Induced Label Propagation (Kernel-LP) framework by mapping is proposed for high-dimensional data classification using the most informative patterns of data in kernel space. The essence of Kernel-LP is to perform joint label propagation and adaptive weight learning in a transformed kernel space. That is, our Kernel-LP changes the task of label propagation from the commonly-used Euclidean space in most existing work to kernel space. The motivation of our Kernel-LP to propagate labels and learn the adaptive weights jointly by the assumption of an inner product space of inputs, i.e., the original linearly inseparable inputs may be mapped to be separable in kernel space. Kernel-LP is based on existing positive and negative LP model, i.e. the effects of negative label information are integrated to improve the label prediction power. Also, Kernel-LP performs adaptive weight construction over the same kernel space, so it can avoid the tricky process of choosing the optimal neighborhood size suffered in traditional criteria. Two novel and efficient out-of-sample approaches for our Kernel-LP to involve new test data are also presented, i.e., (1) direct kernel mapping and (2) kernel mapping-induced label reconstruction, both of which purely depend on the kernel matrix between training set and testing set. Owing to the kernel trick, our algorithms will be applicable to handle the high-dimensional real data. Extensive results on real datasets demonstrate the effectiveness of our approach.

## I. Introduction

In the practical applications, many types of real data, such as images, often contains high-dimensional attributes, redundant information and unfavorable features, thus how to represent and classify the real data automatically by machine learning efficiently and effectively is still a challenging task. Besides, most of real data have no explicit class information and are usually hard to identify due to the high-dimensional features and unfavorable features [1-3]. In other words, a lot of real data are unlabeled, whose labels are needed to be estimated.

- Z. Zhang and L. Jia are with School of Computer Science and Technology, Soochow University, Suzhou 215006, China. (e-mails: cszzhang@gmail.com, 20155227021@stu.suda.edu.cn)
- M. Zhao is with the Department of Electronic Engineering, City University of Hong Kong, Kowloon, Hong Kong. (e-mail: mzhao4@cityu.edu.hk)
- G. Liu is with the School of Information and Control, Nanjing University of Information Science and Technology, Nanjing, China. (e-mail: gcliu@nuist.edu.cn)
- M. Wang is with the School of Computer and Information, Hefei University of Technology, Hefei, China. (e-mail: eric.mengwang@gmail.com)
- S. Yan is with the Department of Electrical and Computer Engineering, National University of Singapore, Singapore. (e-mail: eleyans@nus.edu.sg)

Note that these issues can benefit from the semi-supervised learning (SSL) methods [1-3] that can learn knowledge using both labeled and unlabeled data, and especially by capturing their geometrical structures over a graph [1][35-37][48-51].

Label Propagation (LP), which is one of the most popular graph based SSL algorithms [1][35-37], has aroused much attention the areas of data mining and pattern recognition in recent years because of its effectiveness and efficiency. More specifically, LP has been successfully applied to various real applications, e.g., face recognition and image segmentation. LP is a process of propagating label information of labeled data to the unlabeled data based on their intrinsic geometry relationships, which is mainly performed via trading-off the manifold smoothness term over neighborhood preservation and the label fitness term [3-13], where the label fitness is to measure the predicted soft labels and the initial states, and the manifold smoothness term enables LP to determine the labels of samples by receiving partial information from neighbors.

Generally speaking, existing graph based LP algorithms [3-13][17-24] can be divided into two categories according to whether it can be applied to classify the outside new data, i.e., transductive and inductive models. The transductive methods aim at predicting the unknown labels of samples directly, i.e., it cannot handle those new outside data effectively, while the inductive ones can handle the data inclusion task effectively through label embedding [4][5][6] or label reconstruction [8]. Representative label embedding based inductive algorithms include Flexible Manifold Embedding (FME) [4], Laplacian Linear Discriminant Analysis (LapLDA) [6], Embedded Label Propagation (ELP) [5] and discriminative Sparse FME (SparseFME) [47]. FME, ELP, LapLDA and SparseFME can solve the out-of-sample issue by learning a linear projection explicitly. Different from LapLDA, both FME and ELP add a regressive error term to encode the mismatch between soft labels and embedded features/soft labels by a classifier [4][6]. Note that LapLDA, FME, ELP and SparseFME are all linear projection methods, so the computations for label estimation mainly depend on the dimensionality of data. Thus, LapLDA, FME, ELP and SparseFME may be inefficient for the very high-dimensional case in reality. By comparing with the label embedding based scheme, the reconstruction based inclusion method aims to reconstruct the label of each new data using the soft labels of its neighbors in training set [8]. Thus, the reconstruction scheme has to firstly search the neighbors of each new data prior to label reconstruction, which is time-consuming especially for the large-scale testing set. Besides, no explicit mapping or projection is delivered for inclusion of new data in the label reconstruction based scheme.

Popular transductive LP methods include Gaussian Fields and Harmonic Function (GFHF) [7], Learning with Local and Global Consistency (LLGC) [3], Special Label Propagation (SLP) [10], Linear Neighborhood Propagation (LNP) [8], Class Dissimilarity based LNP (CD-LNP) [9], Nonnegative Projective Label Propagation by Label Embedding (ProjLP) [11], Sparse Neighborhood Propagation (SparseNP) [12], and Positive and Negative Label Propagation (PN-LP) [13], etc. Notice that the existing GFHF, LLGC, SLP, LNP, CD-LNP, ProjLP and SparseNP algorithms mainly discuss the problem that "sample $i$ should be assigned to class $l$" by trading-off the manifold smoothness and label fitness terms. But recent PN-LP extends the common setting by considering additional information that "sample $i$ should not be assigned with the label $k$ $(k{\neq}l)$ jointly", and presents an associated model by including negative label information in the process of label prediction. Note that existing transductive algorithms cannot deal with the outside new data directly and efficiently.

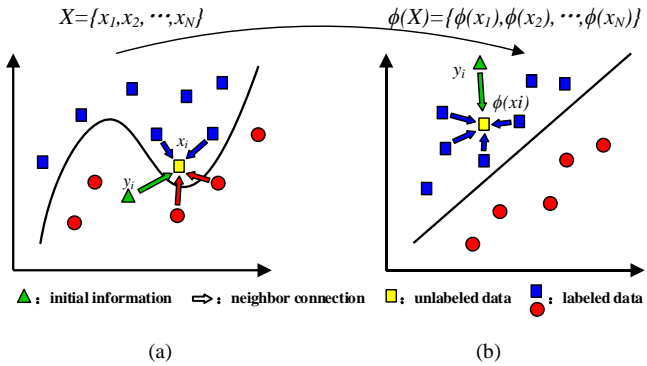

**Fig. 1:** A comparison between label propagation in Euclidean space (a) vs. kernel space (b) for classification.

It should be noticed that the above-mentioned LP methods suffer from the following common drawbacks. First, to the best of our knowledge, the learning process of almost all the existing transductive models and the training process of most inductive models estimate the labels of unlabeled data and compute the graph weights for neighborhood preservation in the original Euclidean space that computes the "ordinary" distance between each data pair, which may be inaccurate in reality, by comparing with the kernel-induced metric over the inner-product distance between samples. In Fig. 1, we show a comparison between label propagation in Euclidean space (a) vs. kernel space (b) for binary-class (denoted by blue square and red circle) data representation and classification, where we also illustrate the neighborhoods of an unlabeled data $x_i$ in Euclidean space and kernel space for visual observation as an example, $y_i$ is the initial label for unlabeled data $x_i$ to receive, and the connections with blue square and red circle denote local neighborhood information to be received by unlabeled data $x_i$. Note that those original linearly inseparable inputs may be mapped into the separable ones in kernel space due to the nonlinear mapping, which is also the main assumption of kernel method [14-15], i.e., the originally linearly inseparable data can be mapped into a higher-dimensional space where it exhibits linear patterns that can be easily represented and classified [14-15]. That is, the patterns in kernel space are more informative and descriptive than the original data in the original Euclidean space. So, the manifold smoothness over the neighborhoods in kernel feature space may be potentially encoded more accurately, since the neighbors of the samples in inseparable Euclidean space may come from other classes, which will lead to inaccurate similarity measure and reduced results directly. Second, for the neighborhood preservation most existing methods pre-compute the weights prior to label propagation by performing the nearest-neighbor-search and weight assignment, so they often suffer from the tricky issue of choosing the optimal neighborhood size. More importantly, the pre-obtained weights using independent step cannot be ensured as optimal for subsequent label prediction. Third, the two existing label embedding and label reconstruction based inclusion methods in testing phase may suffer from the same issue as training process. Specifically, the label embedding methods use the learnt classifier from the Euclidean space for mapping data into label space, while the label reconstruction method also decides the neighbors of new data and assigns weights for the neighbor selection in Euclidean space.

In this paper, to overcome the aforementioned problems of existing LP models, we propose a solution scheme to address the problems and enhance the performance. Specifically, we propose a novel kernelized LP framework to propagate label information in the kernel feature space [14-15] in an adaptive manner. Next, we summarize the main contributions:

(1) Technically, a new Kernel-Induced Label Propagation (Kernel-LP) model is proposed for transductive classification of high-dimensional samples. For transductive learning, our Kernel-LP performs the joint label propagation and adaptive weight learning in a transformed kernel space via an implicit mapping from the original input space. That is, our Kernel-LP changes the task of label prediction from the commonly-used Euclidean space to the kernel space. The motivation of Kernel-LP to propagate label information and compute the graph weights in kernel space arises from the assumption that the original linearly inseparable input data may be mapped to be linearly separable in kernel space [14-15]. Note that we mainly propose a kernelized LP framework and most existing LP methods can be similarly performed in kernel space. In this paper, we mainly formulate our Kernel-LP based on the existing PN-LP model by the positive effects of the negative constraint to enhance the label prediction power. The sparse L2,1-norm is also regularized on the label predictor in the kernel space, which can ensure the prediction results to be robust against noise and make the results more discriminating and accurate potentially [25-26][34].

(2) To ensure the constructed graph weights to be optimal for the joint representation and classification, to enhance the representation power by learning with the most informative patterns of data in the kernel feature space, and to avoid the tricky process of selecting the optimal neighborhood size, our Kernel-LP performs label propagation in an adaptive manner. That is, our Kernel-LP performs the joint label prediction and adaptive weight construction in the same kernel feature space. Specifically, the graph weights are shared in the kernelized data space and kernelized label space over the inner-product distances of data in kernel space jointly. As a result, the label prediction power can be potentially improved due to the joint minimization of the adaptive representation and classification errors in the kernel feature space.

(3) To handle outside new data effectively and efficiently, we also present two novel out-of-sample extension methods for our Kernel-LP, namely, direct kernel mapping and kernel mapping-induced label reconstruction schemes. The direct kernel mapping approach can involve new test data simply by directly computing the kernel vector (i.e., soft label vector of the test data) between training data and new data, which is so easy to implement. The kernel-induced label reconstruction scheme also purely depends on the kernel matrix between the training set and testing set, where the pairwise similarity and predicted soft label vectors are all obtained from the kernel vectors. Compared with the label reconstruction step of LNP, our kernel-induced label reconstruction is more efficient.

The outline of this paper is shown below. Section II briefly reviews the related work and kernel trick. Section III shows the transductive kernel LP framework, and the out-of-sample extensions are shown in Section IV. The relationship analysis is described in Section V. Section VI shows the settings and results. The paper is finally concluded in Section VII.

## II. RELATED WORK

In this section, we briefly review the closely related works to our method, i.e., PN-LP [13] and Kernel method [14-15].

### A. Positive and Negative Label Propagation (PN-LP)

PN-LP extends the existing LP framework to the scenario of label propagation with both positive and negative labels [13]. Given a set of samples $X = \{x_1, x_2, ..., x_N\} \in \mathbb{R}^{n \times N}$ that belong to label set $L = \{1, 2, ..., c\}$, $n$ is the original dimensionality of each sample $x_i$, $N$ is the number of samples and $c$ is the class number. We assume that each sample only belongs to a class. For SSL, $l$ points $x_i$ are considered as labeled, and the rest $u$ samples are regarded as unlabeled. Let $G = (V, E)$ be a graph, where vertices in $V$ correspond to the data points and edges in $E$ characterize pairwise relationships. An edge connecting the nodes $i$ and $j$ is assigned with a weight $W_{ij}$ that indicates the similarity between them. Usually, this similarity weight can be computed by the heat kernel equation:

$$W(x_i, x_j) = W_{ij} = \exp\left(-\|x_i - x_j\|^2 / \sigma\right), \quad (1)$$

where $\sigma$ is the mean edge length distance among neighbors, and $\|x_i - x_j\|$ is the Euclidean distance between $x_i$ and $x_j$. By extending the traditional LP algorithm to incorporate both positive and negative label information in the process of label estimation, PN-LP defines the following objective function:

$$\Omega(F) = \frac{1}{2} tr(F^T L F) \\ + \frac{\mu}{2} \left[ \mu_1 tr\left((F - Y^+)^T (F - Y^+)\right) - \mu_2 tr\left((F - Y^-)^T (F - Y^-)\right) \right], \quad (2)$$

where $F = [f_1, f_2, ..., f_{l+u}] \in \mathbb{R}^{c \times (l+u)}$ is the predicted soft label matrix to be obtained, $L = D^{-1/2} W D^{-1/2}$ is the normalized graph Laplacian [13], $W$ is a weight matrix, $D = diag_i\{\sum_{j=1:N} W_{i,j}\}$ is a diagonal matrix, and the parameters $0 < \mu < 1$, $\mu_1, \mu_2$ are used to balance the contributions of positive label information $Y^+ = [y_1^+, y_2^+, ..., y_{l+u}^+] \in \mathbb{R}^{c \times N}$ and negative label information $Y^- = [y_1^-, y_2^-, ..., y_{l+u}^-] \in \mathbb{R}^{c \times N}$, defined as

$$y_{ij}^+ = \begin{cases} 1, & \text{if from prior knowledge } x_j \in c_i \\ 0, & \text{otherwise} \end{cases}, \\ y_{ij}^- = \begin{cases} 1, & \text{if from prior knowledge } x_j \notin c_i \\ 0, & \text{otherwise}. \end{cases} \quad (3)$$

where $y_i^+$ and $y_i^-$ denote the initialized positive label and negative label for each sample $x_i$. Clearly, the first term in Eq. (2) is the graph regularization, the second term forces the initially labelled data to retain their initial label, and the third term restricts the initially negative labelled data in obtaining the negative label information. The task of positive LP tries to solve the issue of spreading label information from a small set of labeled data to a much larger set of unlabeled samples, while negative LP refers to the dual problem of positive LP, That is, negative LP propagates the labeling constraint that the $i$-th sample does not belong to the $l$-th class, which means that the actual label information in negative LP is unknown and renders the labeling constraint as less informative than positive labels. By setting the partial derivative w.r.t. $F$ to zero, we can obtain the optimal solution of $F$ as

$$F^* = \left[L + (\mu \mu_1 - \mu \mu_2) I\right]^{-1} \left(\mu \mu_1 Y^+ - \mu \mu_2 Y^-\right). \quad (4)$$

Finally, a high value of $F_{i,j}^*$ denotes a high probability that sample $i$ is assigned to class $j$, and the position corresponding to the largest values decides the class assignment of samples.

### B. Kernel Method and Kernel Trick

We also briefly review the kernel method and kernel trick. Kernel methods are successfully applied to the cases that the input-output relationship may not be linear and the inter-class data are not be separable by a linear boundary (i.e., linearly inseparable) [14-15][39-40]. For the linearly inseparable data, kernel method aims at mapping them to higher dimensions where they can exhibit linear patterns and then linear models can be used in the new feature representation space. Denote by $\phi : x \in X \to \phi(x) \in K$ the nonlinear mapping from original space $\mathbb{R}^n$ to a high-dimensional kernel space $\mathbb{R}^p$ ($p \gg n$), which can be implicitly defined by a kernel function, and the then the inner product $\langle \phi(x_i), \phi(x_j) \rangle$ in the kernel space $K$ between $x_i$ and $x_j$ can be defined as

$$\langle \phi(x_i), \phi(x_j) \rangle = \phi(x_i)^T \phi(x_j) = K(x_i, x_j), \quad (5)$$

which is also the pairwise similarity between each pair of samples. There are four popular kernels for real-valued inputs, i.e., Linear Kernel, Quadratic Kernel, Polynomial Kernel and Gaussian Radial Basis Function (RBF) Kernel.

## III. KERNEL-INDUCED LABEL PROPAGATION BY MAPPING FOR SEMI-SUPERVISED CLASSIFICATION

### A. Summary of Main Notations

We first present the important notations used in this paper. Denote $X = [X_L, X_U]$, where $X_L$ is a labeled set and $X_U$ is a unlabeled subset, assume that there are $c$ classes in $X_L$ and each data of $X_L$ has a unique class label in $\{1,...,c\}$. For any matrix $A=[a_1,a_2,...,a_q]\in\mathbb{R}^{m\times q}$, its $l_{r,p}$-norm is defined as

$$\|A\|_{r,p}=\left[\sum_{i=1}^{m}\left(\sum_{j=1}^{q}|A_{i,j}|^r\right)^{p/r}\right]^{1/p}. \quad (6)$$

When $p = r = 2$, it is the commonly used Frobenius norm; when $p =1$ and $r = 2$, it identifies the L2,1-norm [25-26]. For any matrix $A\in\mathbb{R}^{m\times q}$, let $a_i$ denote the $i$-th column vector of $A$, and $a^i$ denote the $i$-th row vector of $A$. Then, the Frobenius norm and L2,1-norm of $A$ can be defined as

$$\|A\|_F^2 = \sum_{i=1}^{m}\sum_{j=1}^{q}A_{ij}^2 = \sum_{i=1}^{m}\|a^i\|_2^2 = tr(A^T A)=tr(AA^T)$$
$$\|A\|_{2,1} = \sum_{i=1}^{m}\left[\sum_{j=1}^{q}A_{ij}^2\right]^{1/2} = \sum_{i=1}^{m}\|a^i\|_2 = 2\,tr(A^T BA) \quad (7)$$

where $B$ is a diagonal matrix with entries being $B_{ii}=1/2\|a^i\|_2$, $i=1,2,...,m$ [25-26], $^T$ is the transpose of a matrix or a vector, $\|\cdot\|_2$ is the only used vector norm (i.e., $\ell_2$-norm), $A_{ij}$ is the $(i, j)$-th entry of the matrix $A$. In addition, $I$ is an identity matrix, and its dimensionality is automatically compatible with the related matrices. For ease of presentation, the horizontal (resp. vertical) concatenation of a collection of matrices or vectors along row (resp. column) is denoted by $[s^1;s^2;...;s^m]$ (resp. $[s_1,s_2,...,s_q]$). Let $K(x,y)=\phi(x)^T\phi(y)$ be the inner product between $x$ and $y$ in kernel space [14-15].

### B. Proposed Formulation

We discuss the problem of propagating positive and negative labels in kernel space in this section. A novel *Kernel Positive and Negative Label Propagation* (Kernel-LP) framework is proposed for adaptive high-dimensional data classification. Note that our Kernel-LP is formulated based on the model of existing PN-LP, so it can inherit all the merits of PN-LP. In addition, our Kernel-LP also improves PN-LP by changing the task of label propagation from the used Euclidean space in PN-LP to a kernel feature space, i.e., performing the label propagation task in a learnt new feature representation space in which the original linearly inseparable data can be mapped to linearly separable patterns in a higher-dimensional kernel space, and then calculating the adaptive graph weights and estimating the unknown labels of samples over the linearly separable patterns, which is relatively easier and will be more accurate compared with handling the original data. Given the original dataset $X$ including both labeled and unlabeled data, rewriting each solution in kernel space as an expansion in terms of the mapped data, each basis projection vector $p$ in the original space can be formulated as $p=\sum_{i=1}^{N}\tau_i\phi(x_i)=\phi(X)\tau$ [14][39]. Let $\phi(X)=(\phi(x_1),...,\phi(x_N))$ and $Q=[\tau_1|...|\tau_d]\in\mathbb{R}^{N\times c}$ be the projection matrix in kernel space, the initial objective function of our Kernel-LP can be defined as

$$\underset{Q,W}{Min}\,J(Q,W)=\sum_{i=1}^{N}u_i^+\|Q^T K(X,x_i)-y_i^+\|-\sum_{i=1}^{N}u_i^-\|Q^T K(X,x_i)-y_i^-\|$$
$$+\alpha\sum_{i=1}^{N}\left(\|\phi(X)-\phi(X)w_i\|+\|Q^T K-Q^T Kw_i\|_F^2+\|w_i\|_2^2\right)+\beta\sum_{i=1}^{N}\|q^i\|_2 \quad, (8)$$
$$s.t.\,w_{ij}\geq 0,\,w_{ii}=0,\,i,j=1,2,...,N$$

where $Q\in\mathbb{R}^{N\times c}$ is a projection to be trained for predicting the soft labels in the kernel space, $K=\phi(X)^T\phi(X)$ is the kernel matrix, the entries $w_{ij}$ in each reconstruction weight vector $w_j$ can measure the similarities between sample pairs, $w_{ij}\geq 0$ is a nonnegative constraint, $w_{ii}=0, i=1,2,...,N$ is included to avoid the trivial solution $W=I$, and $W=[w_{ij}]\in\mathbb{R}^{N\times N}$ is the reconstruction weight matrix. That is, the weight matrix $W$ is also a variable to be optimized, i.e., our Kernel-LP is able to perform the joint label prediction and reconstruction weight construction in an adaptive manner. More importantly, jointly minimizing $\sum_{i=1}^{N}\left(\|\phi(X)-\phi(X)w_i\|+\|Q^T K-Q^T Kw_i\|_F^2+\|w_i\|_2^2\right)$ can ensure the learnt reconstruction weight matrix $w_{ij}$ to be joint-optimal for representation and classification explicitly, where $\sum_{i=1}^{N}\|Q^T K-Q^T Kw_i\|_F^2$ encodes the manifold smoothness in the kernel space. $\alpha$ and $\beta$ are two positive parameters, $u_i^+$ and $u_i^-$ balance positive and negative label information. Note that the learnt weight matrix $W$ in our model is not symmetric.

It is also worth noting that our Kernel-LP can estimate the labels of samples by performing positive and negative label propagation in the kernel feature subspace. More specifically, the joint minimization of the label reconstruction errors $\sum_{i=1}^{N}u_i^+\|Q^T K(X,x_i)-y_i^+\|-\sum_{i=1}^{N}u_i^-\|Q^T K(X,x_i)-y_i^-\|$ can enforce the vector $Q^T K(X,x_i)$ to be the soft label vector in kernel space, where $u_i^+$ and $u_i^-$ are weighting factors for the labeled and unlabeled samples during the process of classification, and $K(X,x_i)$ is the kernel computational result between the original data matrix $X$ and each data vector $x_i$. That is, the soft label vector $Q^T K(X,x_i)$ of each sample $x_i$ can be easily obtained by embedding $K(X,x_i)$ directly onto the projection directions in $Q$, i.e., the projection $Q$ can also be explicitly regarded as a classifier for the class assignments.

By using matrix expressions, the final objective function of our Kernel-LP algorithm can be formulated as

$$\underset{Q,W}{Min}\,J(Q,W)=$$
$$tr(Q^T K-Y^+)U_1(Q^T K-Y^+)^T - tr(Q^T K-Y^-)U_2(Q^T K-Y^-)^T \quad (9)$$
$$+\alpha\left(\|\phi(X)-\phi(X)W\|_F^2+\|Q^T K-Q^T KW\|_F^2+\|W\|_F^2\right)+\beta\|Q\|_{2,1}$$
$$s.t.\,W\geq 0,\,W_{ii}=0$$

where $K=\phi(X)^T\phi(X)$ is a kernel matrix of $N\times N$. Note that the imposed sparse L2,1-norm on $Q$ by the regularization term $\|Q\|_{2,1}=\sum_i\|q^i\|_2$ can ensure $Q$ is sparse in rows so that discriminative representations, that is, predicted soft labels of data points, can be obtained in the kernel space for predicting their labels [25]. The mixed signs or noise in the predicted soft labels can also be potentially reduced due to the positive effects of robust L2,1-norm regularization [50]. $U_1$ and $U_2$ are two diagonal matrices with the entries being $u_i^+$ and $u_i^-$ respectively for weighting the labeled and unlabeled data in the process of positive and negative label propagation. On one hand, for positive label propagation, $u_i^+$ is usually set to a large value (e.g., $+\infty$) for labeled data so that the resulted label can keep consistent with the initial states, and is set to be a small value (e.g., 0) for the unlabeled data so that label information from the neighbors can be received for class assignments [5]. On the other hand, for positive and negative label propagation, the significance of positive and negative labels is usually unequal, and the increased accuracy can be obtained when the significance of positive labels is higher than negative ones [13]. Based on the above analysis, in this work we set $u_i^+ = 10^{10}$ (to approximate $+\infty$) and 0 for labeled and unlabeled data respectively, and set $u_i^- = 1$ and 0 for the labeled and unlabeled data respectively in simulations. Next, we detail the optimization procedures.

*C. Optimization*

In this section, we show the optimization for the objective function of our presented framework in Eq. (9). It is worth noting that our Kernel-LP involves two major variables, i.e., $Q$ and $W$. More importantly, the two variables depend on each other, so the problem cannot be solved directly. Thus, we aim to solve the problem by updating the variables alternately. That is, we update one variable each time by fixing another one. It is worth noting that our formulation can be optimized alternately between the following two steps: kernelized label propagation by feature mapping and kernel-induced adaptive weight construction. We first fix the weight matrix $W$ to date the projection classifier $Q$ for adaptive label prediction.

**(1) Kernelized label propagation by feature mapping for adaptive classification:**

When the adaptive weight matrix $W$ is known, we can focus on estimating the labels of samples by performing adaptive kernel positive and negative label propagation to compute the kernel-induced projection classifier $Q$. By removing terms independent of $Q$ from the problem in Eq. (9), the reduced problem for computing $Q$ can be obtained as

$$\underset{Q}{Min} \, J(Q) = tr\left(Q^T K - Y^+\right) U_1 \left(Q^T K - Y^+\right)^T$$
$$- tr\left(Q^T K - Y^-\right) U_2 \left(Q^T K - Y^-\right)^T \quad (10)$$
$$+ \alpha \left\| Q^T K - Q^T K W \right\|_F^2 + \beta \left\| Q \right\|_{2,1}$$

By the property of L2,1-norm, we have $\|Q\|_{2,1} = tr(2Q^T V Q)$ [25-26], where $V$ is a diagonal matrix with the entries being $v_{ii} = 1/\left(2\|q^i\|_2\right)$ and $q^i$ is the *i*-th row of $Q$. But note that the derivative does not exist when $q^i = 0$, $i = 1, 2, ..., N$. Thus, when each $q^i \neq 0$, the above formulation can be approximated by involving one more variable $V$ similarity as [25-26]:

$$\underset{Q,V}{Min} \, \wp(Q,V) = tr\left(Q^T K - Y^+\right) U_1 \left(Q^T K - Y^+\right)^T$$
$$- tr\left(Q^T K - Y^-\right) U_2 \left(Q^T K - Y^-\right)^T, \quad (11)$$
$$+ \alpha tr\left(Q^T K L K^T Q\right) + \beta tr\left(Q^T V Q\right)$$

since $\left\| Q^T K - Q^T K W \right\|_F^2 = tr\left(Q^T K L K^T Q\right)$, where $L = (I - W)(I - W)^T$ is a symmetric matrix. By taking the derivative of the above problem with respect to variable $Q$, we can obtain

$$\partial \wp(Q,V)/\partial Q = K U_1 K^T Q - K U_1 Y^{+T} - K U_2 K^T Q \quad (12)$$
$$+ K U_2 Y^{-T} + \alpha K L K^T Q + \beta V Q$$

By setting the derivative $\partial \wp(Q,V)/\partial Q = 0$, we can obtain the projection classifier $Q_{t+1}$ at the *(t+1)*-th iteration as

$$Q_{t+1} = \left( K U_1 K^T - K U_2 K^T + \alpha K L_t K^T + \beta V_t \right)^{-1} \quad (13)$$
$$\times \left( K U_1 Y^{+T} - K U_2 Y^{-T} \right)$$

where $L_t = (I - W_t)(I - W_t)^T$. After the classifier $Q$ is obtained, we can update the diagonal matrix $V$ alternately. Specifically, the *i*-th diagonal element $(v_{t+1})_{ii}$ of the matrix $V_{t+1}$ at the *(t+1)*-th iteration can be computed by

$$(v_{t+1})_{ii} = 1/\left[2 \left\| q_{t+1}^i \right\|_2\right], \, i = 1, 2, ..., N, \quad (14)$$

where $q_{t+1}^i$ is the *i*-th row of $Q_{t+1}$. After the projection $Q$ and diagonal matrix $V$ are updated in kernel space, we can compute the weights $W$ by optimizing the following adaptive graph construction process.

**(2) Kernel-induced adaptive weight construction:**

After the projection matrix $Q$ is computed, we can construct the adaptive reconstruction weights $W$ in kernel space easily. Due to the feature mapping from the original input space to a high-dimensional kernel space, the neighborhoods of the data can be potentially discovered more accurately in the obtained new representation space. As a result, the similarity measure based on the kernel induced adaptive weights $W$ will be more accurate potentially and hence enhancing the label prediction power. The main idea of the adaptive weights construction in kernel space is to jointly minimize the reconstruction error $\|\phi(X) - \phi(X)W\|_F^2$ over the new representations and the label reconstruction error $\|Q^T K - Q^T K W\|_F^2$ in kernel feature space, which ensures the learnt adaptive weight matrix $W$ is optimal for the kernel-induced representation and classification at the same time. By removing the terms independent of $W$ from the objective function of Kernel-LP in Eq.(9), the kernel-induced adaptive weight construction can be formulated as

$$\underset{W}{Min} \, J(W) = \alpha \left( \left\| \phi(X) - \phi(X)W \right\|_F^2 + \left\| Q^T K - Q^T K W \right\|_F^2 + \left\| W \right\|_F^2 \right). \quad (15)$$
$$s.t. \, W \geq 0, W_{ii} = 0$$

To facilitate the optimizations, the term $\|\phi(X) - \phi(X)W\|_F^2$ is firstly transformed into the following equation:

$$\left\| \phi(X) - \phi(X)W \right\|_F^2$$
$$= tr\left[\left(\phi(X) - \phi(X)W\right)^T \left(\phi(X) - \phi(X)W\right)\right]. \quad (16)$$
$$= tr\left( K - KW - W^T K + W^T K W \right)$$

Then, by taking the derivative of the optimization problem in Eq. (15) with respect to $W$, we can obtain

$$\partial J(W)/\partial W = KW - K + K^T Q Q^T K W - K^T Q Q^T K + W. \quad (17)$$

By further setting $\partial J(W)/\partial W = 0$, we can update $W_{t+1}$ at the *(t+1)*-th iteration easily as

$$W_{t+1} = \left( K + K^T Q_{t+1} Q_{t+1}^T K + I \right)^{-1} \left( K + K^T Q_{t+1} Q_{t+1}^T K \right). \quad (18)$$
$$W_{t+1} = \max(W_{t+1}, 0), \, (W_{t+1})_{ii} = 0$$

After convergence of the algorithm, we obtain an optimal adaptive reconstruction weight matrix $W^*$ and classifier $Q^*$. Then, we can obtain the estimated soft labels of samples as

**Algorithm 1 : Transductive Kernel-LP Framework**

**Inputs:** Data matrix $X = [x_1,...,x_l, x_{l+1},...,x_{l+u}] \in \mathbb{R}^{n \times N}$, initial class label sets $Y^+$ and $Y^-$, parameters $\alpha$ and $\beta$, $t = 0$, $Q_0 = 0$.

*While not converged do*
1. Fix $W_t$, update the projection classifier $Q_{t+1}$ and diagonal matrix $V_{t+1}$ by Eqs.(13) and (14), respectively;
2. Fix $Q_{t+1}$, update the adaptive weights $W_{t+1}$ by Eq.(18);
3. Convergence check: if $\|Q_{t+1} - Q_t\|_F^2 \le \varepsilon$, stop; else $t = t+1$.

*End while*

**Output:** An optimal projection classifier $Q^* = Q_{t+1}$.

$$F^* = Q^* \phi(X)^T \phi(X) = Q^{*T} K, \quad (19)$$

where the position corresponding to the biggest element in label vector $f_i^*$ determines the class assignment of each $x_i$. That is, the hard label of each data point $x_i$ can be assigned as $\arg\max_{j \le c} f_{ji}^*$, where $f_{ji}^*$ is the $j$-th entry of estimated soft label vector $f_i^*$. For complete presentation of the method, we summarize the optimization procedures of our Kernel-LP in Algorithm 1, where the weight matrix $W_0$ is initialized as $(K + K^T Q Q^T K + I)^{-1}(K + K^T Q Q^T K)$, and the diagonal matrix $V_0$ is initialized as an identity matrix as [25-26].

### D. Computational Time Complexity

We analyze the computational complexities of our Kernel-LP and other traditional methods. Note that GFHF, LLGC, SLP and LNP need to construct the $k$ nearest neighborhood graph, and the computational complexity is $O(N^2 k)$, where $k$ is the number of neighborhoods and $N = l + u$. For label propagation, the computational complexity of GFHF, LNP, LLGC and SLP is $O(Nkc)$, where $c$ is the number of classes. So, the total complexity GFHF, LNP, LLGC and SLP is $O(N^2 k + Nkc)$. Since LNP has to additionally solve the reconstructed weight matrix by solving a standard QP problem, the computational complexity is $O(Nk^3)$. Hence, the computational complexity of LNP is $O(N^2 k + Nkc + Nk^3)$. For our Kernel-LP method, the computational complexity of calculating the projection classifier is $O(Nsc)$, where $s$ is the average number of the nonzero of data points [42]. Since Kernel-LP does not need to construct a $k$ nearest neighborhood graph and compute the adaptive weights, so the computational complexity is $O(N^3)$. The complexity of computing the kernel matrix $K$ is $O(nN^2)$. Since $s$ is at most equal to $e$ and $c \ll N$, if we use a big $O$ to represent the complexity, then the overall complexity of our Kernel-LP will be $O(N^3)$ if $n \le N$, that is, our Kernel-LP also depends more on the number of training data than the input dimensionality as other existing methods.

### E. Convergence Analysis

Note that our proposed Kernel-LP formulation involves three main variables and optimizes the variables in an alternative way, so we would like to analyze its convergence behavior. To assist the proof, a lemma in [25] is firstly reviewed.

**Lemma 1.** For any pair of non-zero vectors $a, b \in \mathbb{R}^m$, the following inequality holds:

$$\|a\|_2 - \|a\|_2^2 / 2\|b\|_2 \le \|b\|_2 - \|b\|_2^2 / 2\|b\|_2. \quad (20)$$

Then, the convergence behavior of our Kernel-LP can be summarized as the following proposition.

**Proposition 1.** The objective function of our Kernel-LP in Eq.(9) is non-increasing at each iteration of the optimizations when the diagonal matrix $V$ is fixed as constant.

*Proof.* Based on the formulations of the sub-problems in Eqs. (10) and (15), since $tr(A^T A) = tr(AA^T)$, the objective function of our Kernel-LP in Eq. (9) can be reformulated as

$$\min_{Q,W,V} tr(Q^T K - Y^+) U_1 (Q^T K - Y^+)^T - tr(Q^T K - Y^-) U_2 (Q^T K - Y^-)^T$$
$$+ \alpha \left( \|\phi(X) - \phi(X)W\|_F^2 + \|Q^T K - Q^T KW\|_F^2 + \|W\|_F^2 \right) + \beta tr(Q^T V Q) \quad (21)$$

Denote $V$ as $V_t$ at the $t$-th iteration, if we aim to calculate $Q_{t+1}$ and $W_{t+1}$ at the $(k+1)$-th times iteration, we can have the following inequality held:

$$\Delta_{t+1} + \alpha \left( \Omega_{t+1} + \|W_{t+1}\|_F^2 \right) + \beta tr(Q_{t+1}^T V_t Q_{t+1})$$
$$\le \Delta_t + \alpha \left( \Omega_t + \|W_t\|_F^2 \right) + \beta tr(Q_t^T V_t Q_t) \quad (22)$$

where the auxiliary matrices $\Delta_{t+1}$ and $\Omega_{t+1}$ are defined as

$$\Delta_{t+1} = tr(Q_{t+1}^T K - Y^+) U_1 (Q_{t+1}^T K - Y^+)^T$$
$$- tr(Q_{t+1}^T K - Y^-) U_2 (Q_{t+1}^T K - Y^-)^T \quad (23)$$

$$\Omega_{t+1} = \|\phi(X) - \phi(X)W_{t+1}\|_F^2 + \|Q_{t+1}^T K - Q_{t+1}^T KW_{t+1}\|_F^2. \quad (24)$$

It should be noted that $\|Q_t\|_{2,1} = \sum_{i=1}^N \|q_t^i\|_2$, so the following equivalent inequality holds for Eq.22:

$$\Delta_{t+1} + \alpha \left( \Omega_{t+1} + \|W_{t+1}\|_F^2 \right) + \beta \left\{ \|Q_{t+1}\|_{2,1} - \left[ \sum_i \|q_{t+1}^i\|_2 - \sum_i \frac{\|q_{t+1}^i\|_2^2}{2\|q_t^i\|_2} \right] \right\}$$
$$\le \Delta_t + \alpha \left( \Omega_t + \|W_t\|_F^2 \right) + \beta \left\{ \|Q_t\|_{2,1} - \left[ \sum_i \|q_t^i\|_2 - \sum_i \frac{\|q_t^i\|_2^2}{2\|q_t^i\|_2} \right] \right\} \quad (25)$$

According to the result in **Lemma** 1, it is easy to obtain the following inequalities:

$$\sum_i \|q_{t+1}^i\|_2 - \sum_i \frac{\|q_{t+1}^i\|_2^2}{2\|q_t^i\|_2} \le \sum_i \|q_t^i\|_2 - \sum_i \frac{\|q_t^i\|_2^2}{2\|q_t^i\|_2}. \quad (26)$$

By combing the above inequalities, we can easily obtain the following inequality:

$$\Delta_{t+1} + \alpha \left( \Omega_{t+1} + \|W_{t+1}\|_F^2 \right) + \beta \|Q_{t+1}\|_{2,1}$$
$$\le \Delta_t + \alpha \left( \Omega_t + \|W_t\|_F^2 \right) + \beta \|Q_t\|_{2,1} \quad (27)$$

which explicitly indicates that the objective function value of Kernel-LP is decreasing monotonically in the optimizations. Note that the above theorem indicates the objective function of our Kernel-LP is non-increasing, but it is also important to show the predicted labels $Q^T K$ by $Q$ also converges, since the projection classifier $Q$ is one major variable. Note that the convergence condition ( $\varepsilon$ is a small value) can be defined as

**Algorithm 2 : Inductive Kernel-LP by Kernel Mapping**

**Inputs:** Training set $X = [x_1,...,x_l, x_{l+1},...,x_{l+u}] \in \mathbb{R}^{n \times N}$ and test set $X_T = \{x_1, x_2,..., x_{N_T}\} \in \mathbb{R}^{n \times N_T}$, initial label sets $Y^+$ and $Y^-$, model parameters $\alpha$ and $\beta$, $t = 0$, $Q_0 = 0$.

**Training Phase based on $X$:**
*While not converged do*
1. Fix $W_t$, update the projection classifier $Q_{t+1}$ and diagonal matrix $V_{t+1}$ by Eqs.(13) and (14), respectively;
2. Fix $Q_{t+1}$ to update the adaptive weights $W_{t+1}$ by Eq.(18);
3. Convergence check: if $\|Q_{t+1} - Q_t\|_F^2 \leq 10^{-5}$, stop; else $t = t+1$.

*End while*

**Testing Phase based on $X$ and $X_T$:**
4. Estimate the soft label matrix of all the testing data by $F_{new} = Q^T K(X, X_T) = [f_1, f_2,..., f_{N_T}] \in \mathbb{R}^{c \times N_T}$, where the position corresponding to the biggest entry in each label vector $f_i$ decides the class assignment of each test data $x_i$.

$$error^{t+1} = \|Q_{t+1} - Q_t\|_F \leq \varepsilon, \quad (28)$$

which can measure the divergence between two sequential projection classifiers and ensure the predicted results will not change drastically, which will be verified by simulations.

## IV. INDUCTIVE EXTENSION OF KERNEL-LP

In the above section, we have introduced the main procedure and transductive formulation of Kernel-LP. In what follows, we will also discuss the issue of extending our Kernel-LP to handle outside new data. More specifically, we will present two approaches for inclusion of the new data by direct kernel mapping and by label reconstruction. For the inductive learning scenario, a test set $X_T = \{x_1, x_2,..., x_{N_T}\} \in \mathbb{R}^{n \times N_T}$ is also available and $X = [X_L, X_U]$ is considered as the training set.

### A. Inclusion of New Data by Direct Kernel Mapping

We first describe the inclusion scheme for our Kernel-LP to include the outside new data through direct kernel mapping, termed I-Kernel-LP-map. According to [8][41], to generalize our Kernel-LP to handle the out-of-sample data, we need to employ the same type of mapping criterion as in Eq.(9) for a new test sample $x_{new}$, and ensure that the inclusion of each sample $x_{new}$ in $X_T$ will not affect the original objective function value on the training dataset $X$. As claimed above, the joint minimization of the label reconstruction errors $\sum_{i=1}^N u_i^+ \|Q^T K(X, x_i) - y_i^+\| - \sum_{i=1}^N u_i^- \|Q^T K(X, x_i) - y_i^-\|$ can enforce the vector $Q^T K(X, x_i)$ to be the soft label vector of training data $x_i$ in kernel space. Thus, the kernel mapping criterion for computing the soft label vector $f_{new}$ and hard label vector $l(x_{new})$ of each new test data $x_{new}$ can be similarly formulated by kernelizing $x_{new}$ with the training data $X$ as

$$\begin{aligned} l(x_{new}) &= \arg\max_{i \leq c} (f_{new})_i, \\ \text{where } f_{new} &= Q^T K(X, x_{new}) \end{aligned} \quad (29)$$

where $(f_{new})_i$ is the $i$-th entry of estimated vector $f_{new}$. That is, the position corresponding to the biggest entry in label vector $f_{new}$ decides the class assignment of $x_{new}$. Note that the soft label matrix $F_{new}$ of all the test data in $X_T$ can be obtained as $F_{new} = Q^T K(X, X_T)$, which is easy and efficient. For complete presentation of the approach, we summarize the inductive Kernel-LP by direct kernel mapping (I-Kernel-LP-map) in Algorithm 2, where the initializations of $W_0$ and $V_0$ are the same as those of transductive Kernel-LP in Algorithm 1.

### B. Inclusion of Outside New Data by Kernel-Induced Label Reconstruction

We also present a kernel-induced label reconstruction scheme for our Kernel-LP model, referred to as Inductive Kernel-LP by kernel-induced label reconstruction (I-Kernel-LP-recons), to involve the new data. This label reconstruction scheme is motivated by the scheme of LNP [8], but it should be noted that our proposed kernel-induced label reconstruction scheme differs from that of LNP clearly in two main aspects. First, the label reconstruction scheme of LNP needs to search the $k$-nearest neighbors of each new data firstly and then computes the weighted soft labels to obtain the label of each new data, while our inductive Kernel-LP-recons only needs to compute the kernel matrix between the training set $X$ and test set $X_T$, which is simple and straightforward. Second, searching the $k$-nearest neighbors and assigning the weights in LNP are all performed in Euclidean space, while our inductive Kernel-LP-recons performs classification clearly in the kernel space. For inductive classification, since $Q^T K(X, x_i)$ is the predicted soft label of each sample $x_i \in X$, we can define the following similar smoothness criterion as Kernel-LP for each $x_{new}$:

$$\vartheta(f_{new}) = \sum_{\substack{i: x_i \in X \\ x_i \in top_k(x_{new})}} \frac{K(x_i, x_{new})}{\|K(X, x_{new})\|_2} (f_{new} - Q^T K(X, x_i))^2, \quad (30)$$

where $K(X, x_{new}) \in \mathbb{R}^N$ is the column kernel vector between training set $X$ and $x_{new}$, and $K(x_i, x_{new})$ is the $i$-th entry of the vector $K(X, x_{new})$. Note that $K(x_i, x_{new})$ can also be treated as the kernel-induced similarity between each training data $x_i$ and $x_{new}$, i.e. $K(x_i, x_{new})/\|K(X, x_{new})\|_2$ is a normalized similarity that can be applied to measure the closeness degree between each $x_i \in X$ and $x_{new}$. Note that $top_k(x_{new})$ contains $k$-closest training samples to sample $x_{new}$ based on sorting $K(X, x_{new})$ in descending order, i.e., the $k$-closest training samples have the largest kernel-induced similarity to the new data $x_{new}$. Since $\vartheta(f)$ is convex in $f_{new}$, it is minimized when

$$f_{new} = \sum_{\substack{i: x_i \in X \\ x_i \in top_k(x_{new})}} \frac{K(x_i, x_{new})}{\|K(X, x_{new})\|_2} Q^T K(X, x_i), \quad (31)$$

which can be easily expressed using the matrix form as

$$f_{new} = Q^T K(X, X_k) \frac{K(X_k, x_{new})}{\|K(X_k, x_{new})\|_2}, \quad (32)$$

where $X_k$ is a data matrix containing the $k$-closest training samples to $x_{new}$, and $K(X, X_k)$ is the kernel matrix between the training set $X$ and $X_k$. It is worth noting that there is no need to compute the soft label matrix $Q^T K(X, X_k)$ and kernel matrix $K(X_k, x_{new})$ separately. Specifically, $Q^T K(X, X_k)$ can be extracted directly from the predicted soft label matrix

$Q^T K(X,X)$ of all training samples, which is achieved in the training phase, based on the indices of the $k$-closest training samples to $x_{new}$ in $top_k(x_{new})$. Similarly, $K(X_k, x_{new})$ can also be obtained directly from $K(X, x_{new})$ based on the indices of the $k$-closest training samples to $x_{new}$. In other words, we only need to compute the kernel computation matrix $K(X, X_T)$ in this reconstruction scheme. Finally, the label of each $x_{new}$ can be optimally reconstructed from the soft labels of the kernel-induced $k$-closest training samples in training set, i.e.,

$$f_{new} = \min_{f_{new}} \left\| f_{new} - Q^T K(X, X_k) \frac{K(X_k, x_{new})}{\|K(X_k, x_{new})\|_2} \right\|^2, \quad (33)$$

where the position corresponding to the biggest entry in label vector $f_{new}$ similarly determines the class assignment of $x_{new}$. We also summarize the procedures of I-Kernel-LP-recons in Algorithm 3, where the initializations of $W_0$ and $V_0$ are the same as those of transductive Kernel-LP in Algorithm 1.

---

**Algorithm 3: Inductive Kernel-LP by Kernel-Induced Label Reconstruction**

**Inputs:** Training set $X = [x_1,...,x_l, x_{l+1},...,x_{l+u}] \in \mathbb{R}^{n \times N}$, test set $X_T = \{x_1, x_2,...,x_{N_T}\} \in \mathbb{R}^{n \times N_T}$, initial class label sets $Y^+$ and $Y^-$, trade-off parameters $\alpha$ and $\beta$, $t=0$, $Q_0=0$.

**Training Phase based on $X$:**
*While not converged do*
1. Fix $W_t$, update the projection classifier $Q_{t+1}$ and diagonal matrix $V_{t+1}$ by Eqs.(13) and (14), respectively;
2. Fix $Q_{t+1}$ to update the adaptive weights $W_{t+1}$ by Eq.(18);
3. Convergence check: if $\|Q_{t+1} - Q_t\|_F^2 \leq 10^{-5}$, stop; else $t = t+1$.

**Testing Phase based on $X$ and $X_T$:**
4. Compute the kernel matrix $K(X, X_T)$ between training set $X$ and test set $X_T$;
5. For each new test data $x_j \in X_T$:
   1) Obtain the similarity vector $K(X, x_j)$ from $K(X, X_T)$;
   2) Obtain the matrix $X_k$ containing $k$-closest training samples to $x_j$ by sorting $K(X, x_{new})$ in descending order;
   3) Extract $Q^T K(X, X_k)$ from the predicted soft label matrix $Q^T K(X, X)$ of all training samples, and extract $K(X_k, x_j)$ from $K(X, x_j)$ based on the indices of the $k$-closest training samples to $x_j$;
   4) Obtain the soft label vector $f_j$ of the test data $x_j$ as $f_j = Q^T K(X, X_k) \overline{w_j}$, where $\overline{w_j} = K(X_k, x_j) / \|K(X_k, x_j)\|$ is the normalized similarity. The biggest element in label vector $f_j$ decides the class assignment of $x_j$.

---

## V. EXPERIMENTAL RESULTS AND ANALYSIS

We mainly evaluate the proposed Kernel-LP model for image classification and segmentation, along with illustrating the comparison results with the related methods for transductive and inductive learning. For transductive classification, the label prediction power of Kernel-LP is mainly compared with those of LNP, SparseNP, CD-LNP, ProjLP, LLGC, GFHF, SLP and PN-LP. For inductive data classification, we mainly evaluate the inclusion performance of our I-Kernel-LP-map and I-Kernel-LP-recons by comparing the results with the two widely-used inductive schemes, i.e., label reconstruction of LNP [8] and the direct label embedding method (including LapLDA, FME, ELP and SparseFME [47]). To avoid the tricky process of choosing different widths $\sigma$ of Gaussian function used in GFHF, LLGC, LapLDA, SLP and FME, the LLE-style reconstruction weights [8][38] are applied in them for semi-supervised learning for fair comparison. CD-LNP and SparseFME use their respective weights. Since the LLE-reconstruction weight matrix $W$ is asymmetric, we preprocess it by symmetrizing it as $W = (W + W^T)/2$ for each algorithm and then normalizing it as $W = D^{-1/2} W D^{-1/2}$, where $D$ denotes a diagonal matrix with the $i$-th entry being the sum of the $i$-th row or column of $W$. The diagonal elements of the weight matrix are also set to be zeros for LLGC. Note that Kernel-LP has two model parameters (i.e., $\alpha$ and $\beta$) to estimate, similarly as FME and ELP. In this simulation, the parameters $\alpha$ and $\beta$ are carefully chosen from the same candidate set for fair comparison, similarly as [18][22][32]. For FME and ELP, the parameters are selected based on the rules that a small $\alpha$ and a relatively large $\beta$ are used according to [4-5]. In all simulations, the neighborhood size $k$ (i.e., number of neighbors) is set to 7 for each method as [43-44] that have shown that this choice can generally work well on the whole.

We mainly evaluate each method by quantitative evaluation of classification and visual observation of segmentation. For fair comparison, all criteria use their respective regularization parameters for weighting the labeled and unlabeled data, and all results are averaged for each setting. The Gaussian kernel, i.e. $K(x_i, x_j) = \exp(-\|x_i - x_j\|^2 / 2\sigma^2)$, is employed in our method, and the width $\sigma$ will be discussed later. In this study, six face databases, three handwriting digit databases, and three object databases, are involved. The experiments are carried out on a PC with Intel (R) Core (TM) i5-4590 @ 3.30Hz 8.00 GB.

### A. Visualization of Graph Adjacency Matrix

Kernel-LP performs joint label prediction and reconstruction weight learning in an adaptive manner. More importantly, jointly minimizing $\sum_{i=1}^{N} \left( \|\phi(X) - \phi(X) w_i\| + \|Q^T K - Q^T K w_i\|_F^2 + \|w_i\|_2^2 \right)$ can ensure the learnt reconstruction weights $W_{ij}$ to be joint-optimal for representation and classification explicitly. Thus, we would like to visualize the learnt graph adjacency matrix and compare the learnt reconstruction weight matrix with the two widely-used methods in the existing label propagation methods to define the weights, i.e., Gaussian function used in GFHF, LLGC, LapLDA, SLP, and the LLE-reconstruction weights are used in LNP. In this study, UMIST face database [2] is used for constructing the weights. UMIST face set has 575 images of 20 persons (mixed race/gender/appearance). Each individual is shown in a range of poses from profile to frontal views. The average size of images is 32×32 pixels. For semi-supervised learning, we choose 10 face images per person as labeled (i.e., 200 images) and treat the rest as unlabeled (i.e., 375 images). Note that the computed LLE-style reconstruction weights, Gaussian weights, and our adaptive reconstruction weights over labeled and unlabeled data are shown in Figs. 2 (a-c) and Figs. 2 (d-f), respectively. From the comparison results, we can see clearly that: (1) The constructed weight matrices by the three kinds of weighting methods have the approximate block-diagonal structures; (2)

---

[2] Available at http://www.sheffield.ac.uk/eee/research/iel/research/face

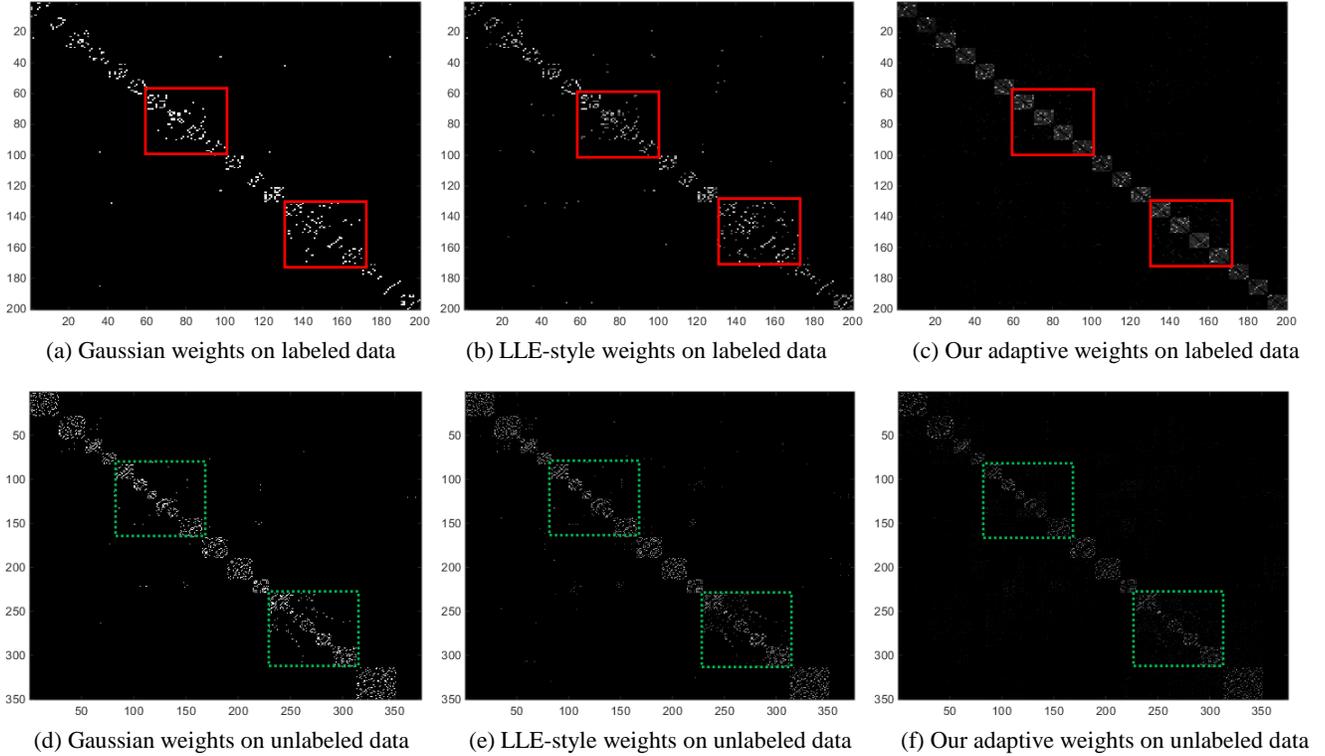

**Fig. 2:** Visualization of the Gaussian weights, LLE-style reconstruction weights, and our adaptive reconstruction weights.

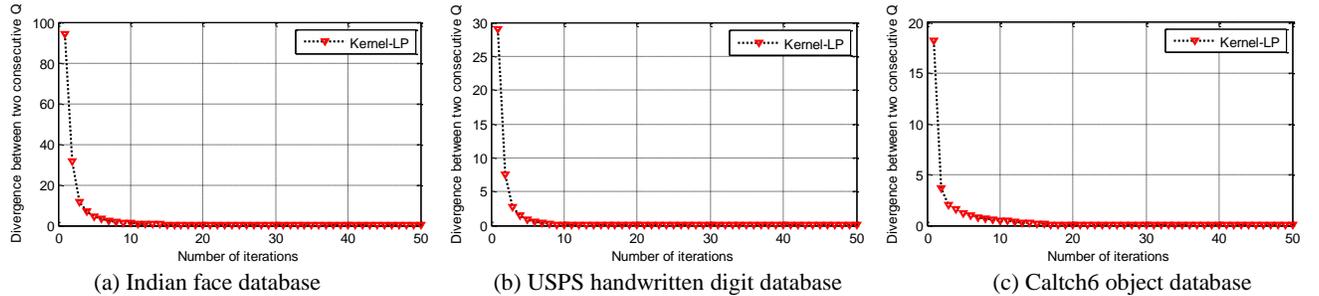

**Fig. 3:** Convergence behavior of our Kernel-LP, where (a-c) show the divergence between two consecutive $Q$ on three databases.

There are more wrong inter-class connections in the weight matrices constructed by Gaussian function and the LLE-style reconstruction weights, for example the highlighted parts by the red and green rectangles, which will result in inaccurate similarity measure, representations and label predictions. In contrast, the adaptive reconstruction matrix in our Kernel-LP is more informative than both the LLE-style reconstruction weights and Gaussian weights, because our adaptive graph weight matrix has less wrong inter-face-images connections and better intra-class connectivity simultaneously. Note that a good intra-class connectivity with less inaccurate inter-class connections is very important for subsequent representation and classification, since the target data can be reconstructed by the samples from the same class as much as possible. It is also noted that the superiority of our learnt adaptive graph adjacency matrix over other two can be attributed to the joint learning process of weights by minimizing the reconstruction error $\|\phi(X) - \phi(X)w_i\|$ and the label reconstruction error $\|Q^T K - Q^T K w_i\|_F^2$ simultaneously, because the pre-known class information of labeled images and the pairwise relationship between both labeled and unlabeled images can enforce the learnt weight matrix to be block-diagonal.

*B. Convergence Analysis*

We have shown that the objective function value of Kernel-LP in Eq.(9) is non-increasing under the updating rules, so we want to describe some quantitative convergence analysis results by reporting the divergence between two sequential projection classifiers $Q$. In this simulation, three kinds of real databases i.e., Indian face [33], USPS handwritten digits [46] and Caltech6 object, are tested. The used USPS dataset has 11,000 examples of handwritten digits, publicly available at http://www.cs.nyu.edu/~roweis/data.html/USPSHandwritten Digits. There are 8-bit grayscale digit images of '0' through '9' and 1100 images of each digit. In the database, the size of each digit image is 16×16 pixels. The Caltech6 object image database contains 3738 images of six object categories and one background category. For each database, we randomly choose 6 samples as labeled, and the convergence analysis results averaged over 15 iterations are illustrated in Fig. 3. As can be observed from the results, the divergence between consecutive projections $Q$ also converges to zero, which means that the final results will not be changed drastically. It is also noticed that the convergence speed is fast, and the number of iterations is about 15-20 in the simulations.

*C. Image Recognition by Transductive Classification*

**(1) Face Recognition**. We firstly examine our Kernel-LP for transductive learning and recognition. The results are mainly compared with those of LNP, SparseNP, CD-LNP, ProjLP, LLGC, GFHF, SLP, LapLDA and PN-LP. In this study, six real face databases, i.e., AR-male and AR-female sample sets [30], ORL face database, UMIST face database [28], Georgia Tech database and Indian face database, are tested. ORL face database has 40 distinct subjects and each has 10 face images. Georgia Tech face database has 50 distinct subjects and each has 15 face images. There are 7 different images of each of 40 distinct subjects in the Indian database. We also merge the face images of Georgia Tech and Indian into a single one called Georgia-Indian for face recognition. For each database, all the face images are resized into $32 \times 32$ pixels, so each face image corresponds to a data point in a 1024-dimensional space. In this study, we evaluate the learning performance of each method by varying the number of labeled face images of each subject. For each fixed number of labeled face images, the results are averaged over 15 times runs. Fig. 4 illustrates the averaged classification results of each algorithm. We can observe that: (1) The performance of each method can be improved by increasing the number of labeled face images of each subject; (2) Kernel-LP can obtain higher classification accuracy compared with the other recent label propagation methods in most cases, especially on the AR-male and AR-female face datasets. One main reason can be attributed to the formulation of transforming the label prediction and weight construction processes from original input space into a kernel space that can discover the relations between images more accurately. The other reason is the adaptive reconstruction weight construction is incorporated into the label propagation process for joint optimization in the kernel space, which can ensure the constructed weights are joint-optimal for similarity measure and classification. The results of GFHF, LapLDA and SparseNP are generally better than those of SLP, ProjLP PN-LP and CD-LNP in most cases. In Table 1, we describe the statistics of each algorithm according to the numerical results in Fig. 4, where we report the mean accuracy, highest accuracy and mean running time for each method. Note that we report the computational time of one iteration for iterative methods. The highest records are highlighted in bold. We can obtain similar findings from the numerical results shown in Table 1, that is, the performance superiority of evaluated methods keeps consistent with the experimental results in Fig. 4. That is, our proposed two inductive methods i.e., I-Kernel-recons and I-Kernel-map deliver better results than the label reconstruction and label embedding based inductive schemes.

**(2) Object Recognition.** We also evaluate our Kernel-LP to recognize the objects of the ETH80 database [31]. ETH80 database contains 8 big object categories with 10 small object categories per big category. Each small object category has 41 images that vary in shape and texture but otherwise appear on a uniform background and roughly share the same size. In this study, we perform classification on the 8 big categories, i.e., one 8-class classification problem is created and tested. Four experimental settings over various numbers of labeled images (i.e., 20, 80, 140 and 200) are randomly selected from each object class for recognition. The results are averaged over the first 15 best records based on 20 realizations of training and test sets. We describe the classification results in Table 2, including the mean accuracy, standard deviation the best accuracy and mean running time of each algorithm in each setting, in which the simulation settings are also shown. We have the following observations: (1) The performance of each method goes up with the increasing numbers of training samples. (2) Kernel-LP delivers higher accuracy than other techniques in most cases, which can also be attributed to the positive effects of negative label information, kernel label propagation, and the seamless integration of adaptive weight construction and kernelized label propagation. SparseNP also performs well by delivering better results than other methods.

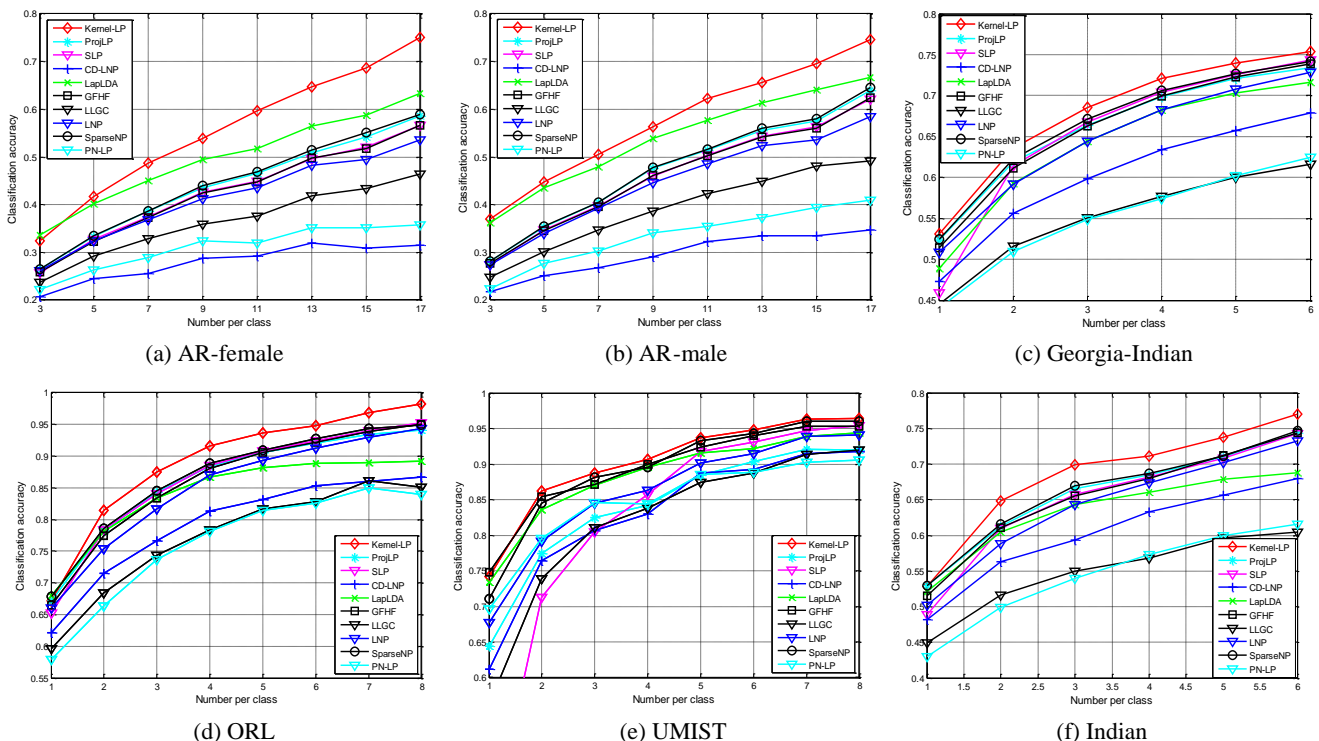

**Fig. 4:** Classification results of each evaluated algorithm on the six face image databases.

**Table 1:** Performance comparison of each algorithm under different settings based on the six face image databases.

| Dataset / Method | AR-female Mean/Best/Time | AR-male Mean/Best/Time | Georgia-Indian Mean/Best/Time | ORL Mean/Best/Time | UMIST Mean/Best/Time | Indian Mean/Best/Time |
|---|---|---|---|---|---|---|
| SparseNP | 26.54/58.85/0.4117 | 27.97/64.50/0.4536 | 66.53/74.112/0.9039 | 86.61/94.93/0.8619 | 89.12/96.05/0.8555 | 65.05/71.51/0.6484 |
| ProjLP | 25.78/58.55/0.3923 | 27.22/63.90/0.4343 | 65.23/73.41/0.8712 | 86.05/94.10/0.8415 | 81.33/92.00/0.8239 | 64.29/71.32/0.6262 |
| SLP | 26.56/56.70/0.4015 | 27.90/62.15/0.4476 | 66.01/74.33/0.8835 | 86.05/95.21/0.8503 | 83.90/95.45/0.8228 | 64.62/71.90/0.6411 |
| LNP | 26.14/53.35/0.3260 | 27.32/58.45/0.3527 | 64.38/72.84/0.8494 | 84.76/94.38/0.7747 | 85.89/94.10/0.8235 | 63.48/71.41/0.5441 |
| LLGC | 22.95/46.45/0.4718 | 24.63/49.20/0.5219 | 55.07/61.57/0.9237 | 77.07/85.14/0.8513 | 81.82/92.00/0.8513 | 53.60/59.42/0.6228 |
| LapLDA | 35.02/63.30/0.3993 | 38.45/66.70/0.4470 | 63.80/71.64/0.9131 | 83.91/89.17/0.8524 | 88.20/94.41/0.8717 | 63.04/67.90/0.6439 |
| GFHF | 26.03/56.70/0.2660 | 27.61/62.35/0.2879 | 65.81/73.85/0.8259 | 85.88/95.07/0.7819 | 89.28/95.35/0.8114 | 64.31/71.80/0.5905 |
| CD-LNP | 20.05/31.55/0.3033 | 20.75/34.75/0.3196 | 59.95/67.85/0.9225 | 79.11/86.74/0.7920 | 82.91/91.76/0.8469 | 59.35/66.53/0.5357 |
| PN-LP | 21.49/35.80/0.4146 | 22.64/40.90/0.4558 | 54.99/62.47/0.9146 | 76.13/83.96/0.8639 | 84.56/90.55/0.8625 | 53.95/61.29/0.6536 |
| **Kernel-LP** | **53.08/75.00**/0.5873 | **53.24/74.55**/0.6033 | **67.76/75.39**/0.8515 | **88.88/98.19**/0.8818 | **90.13/96.46**/0.8152 | **66.46/73.28**/0.6692 |

**Table 2:** Performance comparison of each algorithm under different settings based on the ETH80 database.

| Result / Method | ETH80 database (20 labeled per class) Mean / Std / Best / Time | ETH80 database (80 labeled per class) Mean / Std / Best / Time | ETH80 database (140 labeled per class) Mean / Std / Best / Time | ETH80 database (200 labeled per class) Mean / Std / Best / Time |
|---|---|---|---|---|
| SparseNP | 71.65/3.61/71.48/173.4 | 83.50/2.76/83.90/173.6 | 85.76/3.03/87.43/173.1 | 86.90/2.52/89.01/0.6134 |
| ProjLP | 67.00/0.61/68.01/32.74 | 81.86/0.59/81.97/16.42 | 84.52/1.63/85.23/12.03 | 85.25/3.52/87.71/0.6442 |
| SLP | 62.62/3.21/62.96/11.31 | 82.95/3.05/83.22/11.55 | 85.48/2.47/87.39/11.42 | 86.96/2.71/88.89/0.3791 |
| LNP | 67.50/2.49/68.50/35.41 | 80.24/2.52/81.79/35.77 | 83.94/2.21/85.84/35.70 | 86.05/2.27/89.80/0.6442 |
| LLGC | 60.49/3.11/61.70/44.44 | 73.21/2.95/75.97/44.90 | 75.76/2.84/76.62/44.81 | 76.43/2.46/78.27/0.5543 |
| LapLDA | 59.90/1.53/60.69/11.17 | 68.70/2.82/70.25/11.09 | 69.61/1.21/69.83/11.08 | 70.90/1.73/7390/0.6850 |
| GFHF | 71.10/5.19/72.16/13.74 | 82.32/2.26/82.51/13.67 | 85.15/3.12/86.39/13.64 | 86.55/2.96/88.00/0.6508 |
| CD-LNP | 67.81/2.68/70.85/9.157 | 79.47/2.53/81.47/8.355 | 82.79/1.92/84.21/8.120 | 85.16/1.95/86.42/0.5066 |
| PN-LP | 57.55/3.02/59.32/10.35 | 70.91/1.26/71.30/10.27 | 74.38/2.65/75.96/10.40 | 77.70/1.80/78.18/0.6859 |
| **Kernel-LP** | **72.85**/3.59/73.58/22.11 | **83.54**/2.73/84.64/22.49 | **87.73**/2.33/88.39/22.19 | **89.11**/2.62/90.17/0.6665 |

**Table 3:** Descriptions of the used real-world databases.

| Dataset Name | Data Type | #Class ($c$) | #Dim ($n$) | #Points |
|---|---|---|---|---|
| UMIST | face | 20 | 1024 | 575 |
| Georgia Tech | face | 50 | 1024 | 750 |
| YALE-Indian | face | 76 | 1024 | 842 |
| Pen-RHD | handwriting | 10 | 16 | 10992 |
| HWDB1.1-D | handwriting | 10 | 196 | 2381 |
| USPS | handwriting | 10 | 256 | 11000 |
| COIL100 | object | 100 | 1024 | 7200 |
| ETH80 | object | 80 | 1024 | 3280 |
| Caltech6 | object | 7 | 1024 | 3738 |

*D. Image Recognition by Inductive Classification*

In this study, we mainly evaluate the inclusion performance of our I-Kernel-LP-map and I-Kernel-LP-recons to handle new data. We mainly compare the inductive classification results with the existing two widely-used inductive schemes, that is, label reconstruction of LNP and direct label embedding. For the label embedding based methods, we include the inductive LapLDA, FME, ELP and SparseFME for comparison. In this study, three kinds of image databases are evaluated, including three face databases, three handwriting digit databases and three object image databases. The detailed information of the evaluated databases is described in Table 3. For inductive data classification, we randomly split each dataset into training and test sets, where training set includes the labeled and unlabeled data, and all samples in test set are unlabeled. The accuracy is gained by comparing the predicted labels with the ground-truth labels provided by the original data corpus.

**(1) Face recognition results.** We first evaluate each model for inductive face recognition. Three face databases, including the UMIST database [28], YALE-Indian dataset and Georgia Tech face database [32], are involved. The YALE-Indian face dataset is a merged set of both YALE [3] and Indian face images. In this study, we mainly report the inclusion results of each method on the testing set. The recognition results averaged 15 random splits of training and testing samples are summarized in Fig. 5, where the horizontal axis is the number of labeled training data in each class, and the vertical axis is the averaged inductive accuracy of each semi-supervised learning method. Clearly, the prediction performance of each method can be effectively improved with the increasing number of labeled training data from each class. It can also be noted that our two inductive methods, i.e., I-Kernel-recons and I-Kernel-map, can deliver better results than the label reconstruction and label embedding based inductive schemes. The label reconstruction based LNP outperforms the other remaining methods on the YALE-Indian and Georgia Tech face datasets, while the label embedding based LapLDA, FME, ELP and SparseFME gains better results than the inductive LNP. For the label embedding based methods, SparseFME can outperform LapLDA, FME and ELP for face recognition over each database.

---
[3] Available at http://vision.ucsd.edu/content/yale-face-database

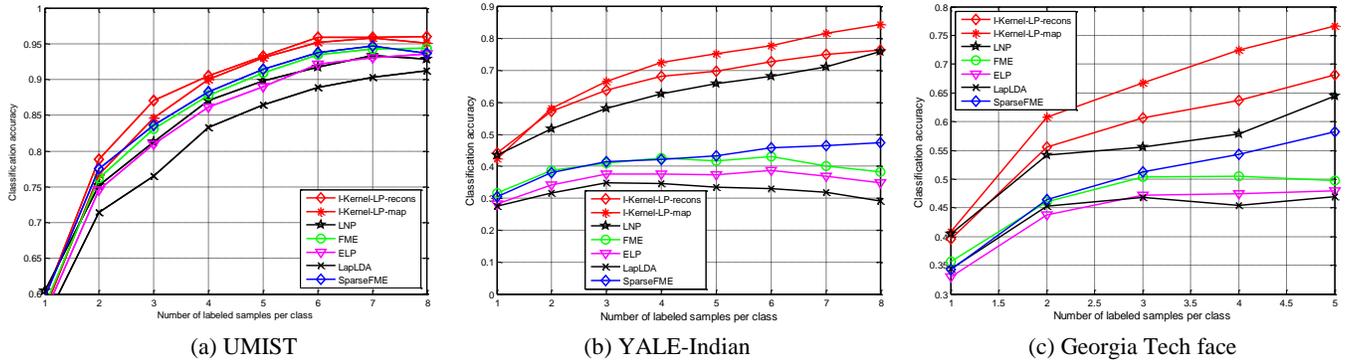

(a) UMIST  (b) YALE-Indian  (c) Georgia Tech face

**Fig. 5:** Inductive classification results of each evaluated algorithm on the three face databases.

**Table 4:** Performance comparison of each algorithm under different settings based on the three face databases.

| Result<br>Method | UMIST face<br>Mean/Best/Time | YALE-Indian face<br>Mean/Best/Time | Georgia Tech face<br>Mean/Best/Time |
|---|---|---|---|
| LapLDA | 0.8052/91.24/0.0019 | 43.76/46.89/0.0023 | 32.00/34.60/0.0011 |
| ELP | 0.8333/93.58/0.0018 | 43.88/47.93/0.0023 | 35.69/37.63/0.0011 |
| FME | 0.8481/94.38/0.0018 | 46.46/49.74/0.0024 | 39.68/42.95/0.0010 |
| SparseFME | 0.8529/93.63/0.0017 | 48.91/58.29/0.0025 | 41.90/47.30/0.0010 |
| Inductive LNP | 0.8396/92.83/0.0805 | 54.51/64.51/0.0574 | 62.17/76.00/0.0311 |
| **I-Kernel-map** | **0.8601/95.12**/0.0204 | **63.51/76.68**/0.0188 | **69.85/84.40**/0.0148 |
| **I-Kernel-recons** | **0.8709/96.01**/0.0365 | **57.49/68.13**/0.0317 | **65.91/76.40**/0.0238 |

Table 4 describes the statistics of each evaluated method according to the quantitative results in Fig. 5, where we report the mean inductive accuracy, best records and inclusion method. The highest two records in each group are highlighted in bold. We find that the performance superiority of evaluated methods can keep consistent with the quantitative analysis in Fig. 5. That is, our two inductive algorithms can obtain higher accuracies compared with the other recent related methods in most cases. SparseFME is capable of achieving better result than the LapLDA, FME and ELP methods. For the running time performance, the label embedding method is the most efficient. The direct kernel mapping and the kernel-induced label reconstruction schemes are more efficient than the label reconstruction method of LNP, because LNP has to search the neighbors of each test data and then computes the weight vector so that the label of test data can be reconstructed by the soft labels of its neighbors, while our proposed data inclusion schemes only depends on the kernel matrix between training and test data. Relatively, the kernel mapping method is more efficient the kernel-induced label reconstruction.

(**2**) **Handwritten digit recognition.** We then evaluate each inductive learning method for handwriting recognition. In this study, three popular handwritten digit databases, including USPS [46], CASIA-HWDB1.1 database [32] and Pen based Recognition of Handwritten Digits [45] (Pen-RHD), are tested. CASIA-HWDB1.1 database has 3755 Chinese characters and 171 alphanumeric and symbols, collected from 300 writers. In this study, the sample set termed HWDB1.1-D [33], including 2381 handwritten digits ('0'-'9'), from CASIA-HWDB1.1 is used for the evaluations. Pen-RHD database contains 10992 handwritten digits of '0'-'9'.

For each handwriting database, we report the results under various numbers of labeled handwritten digits, fix the number of the unlabeled digits to 10, and regard the rest ones as testing set for inclusion. The handwritten digit recognition accuracies of the testing data averaged over 15 times runs are illustrated in Fig. 6. From the results, we can similarly conclude that the overall performance of each method can be greatly improved with the increasing numbers of labeled handwritten digits. More specifically, the performance of our inductive methods and the label reconstruction based LNP goes up faster when the labeled number increases in each setting. We also see that our I-Kernel-map can achieve more promising results than the other recent methods on each database. Kernel-recons obtains highly comparable results with the label reconstruction based LNP on the USPS database, and both are superior to the other remaining methods. In other words, the label reconstruction scheme of LNP and our inclusion schemes of kernel-induced mapping/label reconstruction are all better than the direct label embedding scheme for the handwriting digits recognition.

Table 5 summaries the results (including the mean, highest accuracies and the inclusion time of testing digits) according to Fig. 6. From the results, similar performance superiority of evaluated methods can be obtained. (1) The mean and highest records of our inductive methods and the label reconstruction based LNP are better than the other methods in most cases. (2) Our proposed I-Kernel-LP-map gains higher accuracies than its competitors in each case, specifically when the number of labeled training digits is relatively small. (3) Considering the running time performance of data inclusion, it is clear that the label embedding scheme by direct projection is still efficient, successively followed by our proposed kernel mapping and kernel-induced label reconstruction schemes. For the label reconstruction scheme, I-Kernel-LP-map is faster than LNP.

**(3) Object recognition.** Finally, we evaluate each inductive learning method for recognizing objects. In this study, three popular object databases, i.e., COIL100 [31], ETH80 [31] and Caltech6, are involved. The COIL100 database contains 7,200 images of 100 objects. The objects were placed on a motorized turntable against a black background. The turn-table was rotated through 360 degrees to vary object pose with respect to a fixed color camera. Images of the objects were taken at pose intervals of five degrees, corresponding to 72 different poses per object. For ETH80 database, we regard each subcategory as a single class, namely, an 80-class classification problem is created and evaluated. We also evaluate the data inclusion performance of each learning method by varying the number of labeled object images in each class. Specifically, we tune the number of labeled training samples from {10, 20, …, 100} for both ETH80 and Caltech6, and vary the number of labeled training samples from {2, 5,…,23} for COIL100. Since we mainly focus on evaluating the data inclusion performance, the number of unlabeled training samples is always fixed to 10 for each experimental setting. We illustrate the averaged result of each algorithm over 10 random splits of training and test samples in Fig.7, and Table 6 summarizes the statistics in the figures according, including the mean accuracy, best records and data inclusion time of testing samples. We can have the following observations. First, the increasing number of labeled training samples can effectively boost the object recognition performance of each method. Second, I-Kernel-LP-map and I-Kernel-LP-recons outperform others on ETH80 and COIL100. I-Kernel-LP-map is able to deliver better results on Caltech6 database by achieving better results. The results of the label reconstruction based LNP and I-Kernel-LP-recons are highly comparative with each other. The performances of the label embedding based methods are close with each other in each case. Third, we once again observe from the inclusion time of testing data that the label embedding based inclusion scheme is very efficient. Our proposed direct kernel mapping based inclusion scheme is also relatively more efficient than others, and the label reconstruction scheme of LNP is relatively slow.

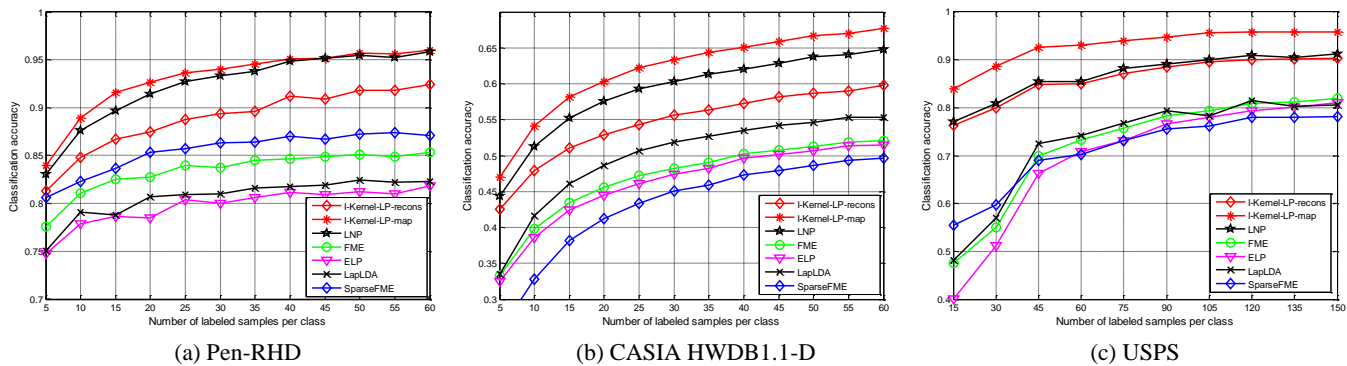

(a) Pen-RHD  (b) CASIA HWDB1.1-D  (c) USPS

**Fig. 6:** Classification results of each evaluated algorithm on the three handwriting digit databases.

**Table 5:** Performance comparison of each algorithm under different settings based on the three handwriting digit databases.

| Result<br>Method | Pen-RHD digits<br>Mean/Best/Time | HWDB1.1-D digits<br>Mean/Best/Time | USPS digits<br>Mean/Best/Time |
|---|---|---|---|
| LapLDA | 80.65/82.47/0.0007 | 49.84/55.32/0.0099 | 72.88/80.66/0.0045 |
| ELP | 79.75/81.86/0.0008 | 46.09/51.52/0.0119 | 69.67/81.02/0.0043 |
| FME | 83.43/85.34/0.0013 | 46.90/52.03/0.0146 | 72.30/81.98/0.0053 |
| SparseFME | 85.49/87.41/0.0012 | 42.91/49.64/0.0145 | 71.36/78.08/0.0053 |
| LNP | **92.35**/95.88/0.5652 | **58.91**/64.80/1.1994 | **86.87/91.24**/0.8293 |
| **I-Kernel-map** | **93.07**/95.59/0.1019 | **61.83**/67.76/0.6987 | **93.01/95.87**/0.3029 |
| **I-Kernel-recons** | 88.85/92.44/0.4160 | 54.45/59.76/1.8411 | **86.18/90.30**/0.8216 |

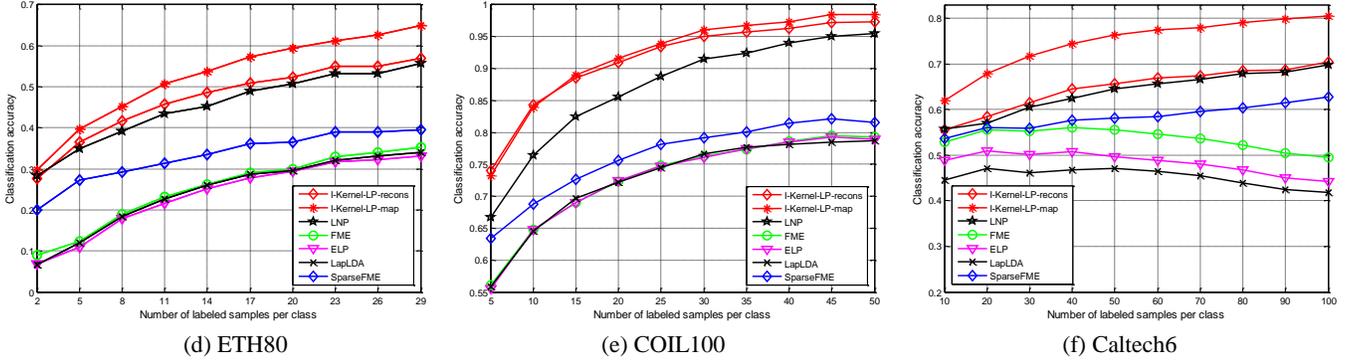

(d) ETH80  (e) COIL100  (f) Caltech6

**Fig. 7:** Classification results of each evaluated algorithm on the three object image databases.

**Table 6:** Performance comparison of each algorithm under different settings based on the three object databases.

| Method \ Result | ETH80 object Mean/Best/Time | COIL100 object Mean/Best/Time | Caltech6 object Mean/Best/Time |
|---|---|---|---|
| LapLDA | 25.24/34.31/0.0070 | 63.75/78.65/0.0195 | 45.18/47.17/0.0055 |
| ELP | 24.66/34.50/0.0076 | 63.76/79.28/0.0223 | 48.33/50.70/0.0056 |
| FME | 26.08/36.11/0.0092 | 65.05/79.46/0.0270 | 53.57/56.13/0.0059 |
| SparseFME | 33.86/40.89/0.0081 | 68.89/82.15/0.0261 | 58.40/62.72/0.0056 |
| Inductive LNP | 46.31/56.67/0.2283 | 76.60/95.44/0.8364 | **63.85/69.85**/0.3770 |
| **I-Kernel-map** | **53.65/65.47**/0.1748 | **83.68/98.50**/0.6931 | **74.76/80.53**/0.1190 |
| **I-Kernel-recons** | **48.08/58.53**/0.2920 | **82.88/97.27**/1.2198 | **64.78/70.46**/0.2151 |

### E. Application to Interactive Image Segmentation

In this section, we prepare a study to interactively segment the natural images using the Berkeley segmentation database [27]. The main task is to extract the foreground regions from natural images. When handling the interactive image segmentation, the most important issue is to collect the user specified pixels about both foreground and background. In this study, seven images from the Berkeley database are tested. Each extracted pixel from the image is represented by using a 5D vector $\eta$, i.e., $\eta = [R,G,B,\lambda,\sigma]^T$, where $(R,G,B)$ denotes the normalized color of pixels and $(\lambda,\sigma)$ denotes the spatial coordinate with image width and height. Fig. 8 exhibits the interactive image segmentation results. The first row shows the original images, and the second row shows the source images with the user specified pixels. Note that the colored lines represent different data, where the red lines represent the foreground that needs to be extracted, referred to one class of labeled pixels, the green lines denote background, regarded as another class of labeled pixels, and the blue lines represent the training pixels without labels. The rest rows illustrate the result of each method.

In this study, we mainly compare the result of Kernel-LP with those of LNP, LapLDA, SLP, LLGC, GFHF and CD-LNP. From the segmentation results in Fig. 8, we can have the following conclusions: 1) Under the same setting, our Kernel-LP generally outperforms other evaluated methods, especially on the castle image, lake Image and swan image, since we can clearly see that more pixels of the foreground and background of images are incorrectly classified by other related methods. In contrast, our Kernel-LP performs better in classifying pixels and output satisfactory results; 2) Detecting the edge regions is relatively difficult. Kernel-LP can deliver visually comparable and even better performance than the others in most cases on determining the image boundaries, since compared methods often make the border of foreground and background blurry and hazy if the colors of foreground and background of target images are similar. LapLDA also gains a better segmentation result, such as on handling the second tea related image.

### F. Parameters Analysis

In this section, we perform two simulations to investigate the effects of model parameters on the learning performing. In the first study, we mainly we investigate the effects of parameters (i.e. $\alpha$, $\beta$) on the performance of our Kernel-LP approach. In another study, we mainly explore the effects of kernel width $\sigma$ in the Gaussian kernel function.

**(1) Investigation of the parameters $\alpha$ and $\beta$.** In this study, we illustrate the results by using the approach of grid search to explore the effects. For each pair of parameters, we average the results over 20 times random splits with varied parameters $\alpha$ and $\beta$ from $\{10^{-5}, 10^{-4},...,10^{4},10^{5}\}$. We choose three kinds of databases (i.e., UMIST face, Pen-RHD digits and COIL100 object) as examples. For the simulations, we fix the number of labeled training images (i.e., 7, 10 and 10 for UMIST face, Pen-RHD digits and COIL100 object databases respectively) and use the remaining samples as unlabeled training set. Three groups of parameter selection results on the three databases are shown in Fig. 9. We can observe from the results that our Kernel-LP can perform well in a wide range of parameters in each group, that is, our proposed method is not sensitive to the parameters $\alpha$ and $\beta$. We also notice that a relatively small $\alpha$ tends to decrease the results on for UMIST face and COIL100 object databases. Similar findings of parameter selections can be found from other sets in most cases, but the results will not be shown due to page limitation. From this study, we simply set $\alpha=10^3$ and $\beta=10^{-1}$ for Kernel-LP in all experiments.

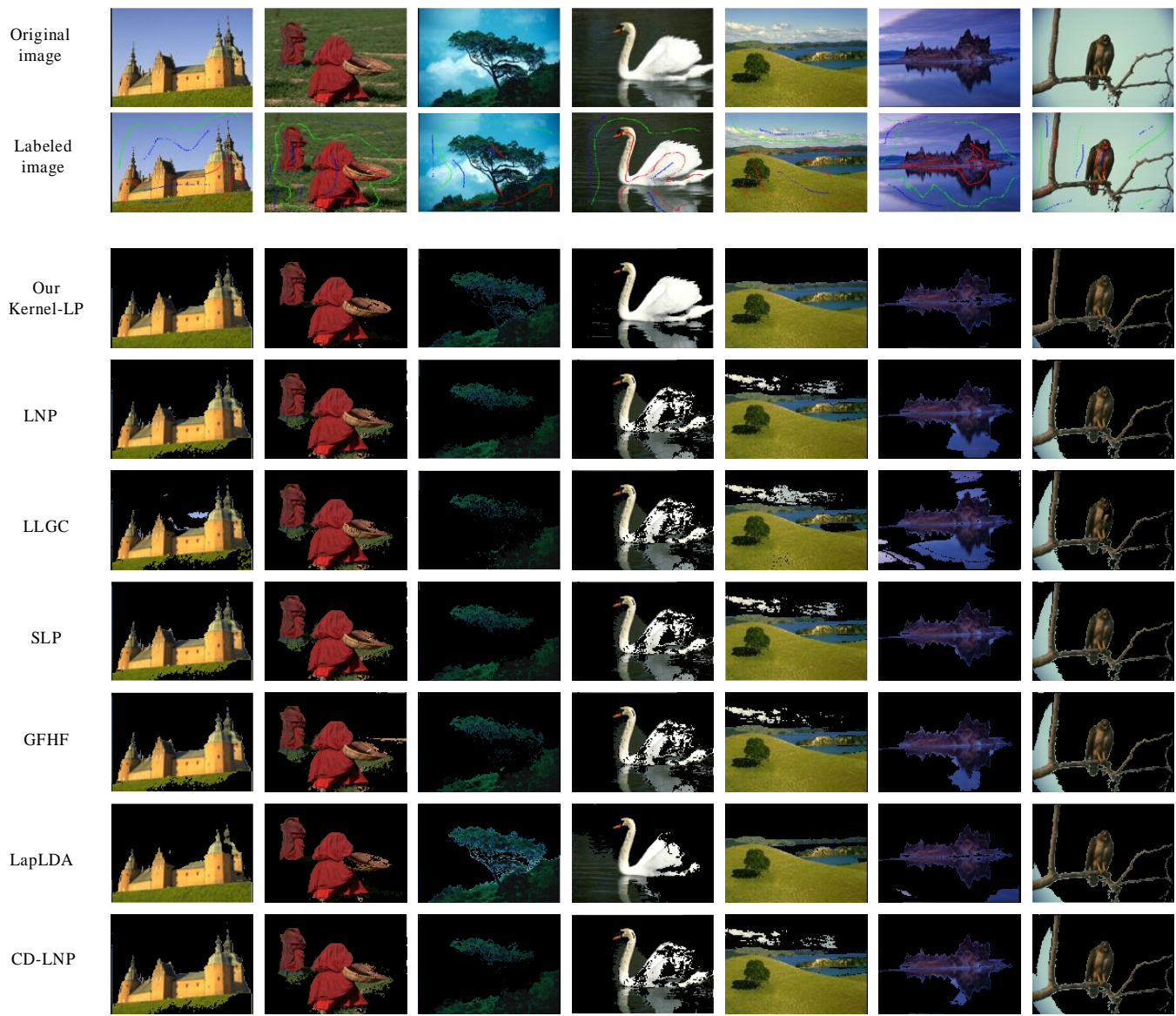

**Fig. 8:** Interactive image segmentation results of each algorithm on the benchmark Berkeley segmentation database.

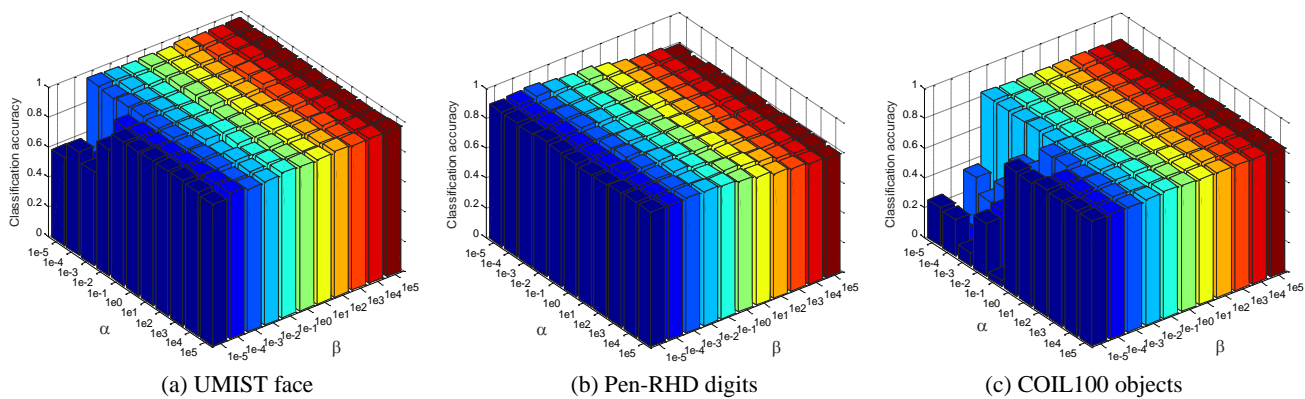

(a) UMIST face  (b) Pen-RHD digits  (c) COIL100 objects

**Fig. 9:** Sensitivity analysis of different selections of parameters $\alpha$ and $\beta$ on three real-world databases.

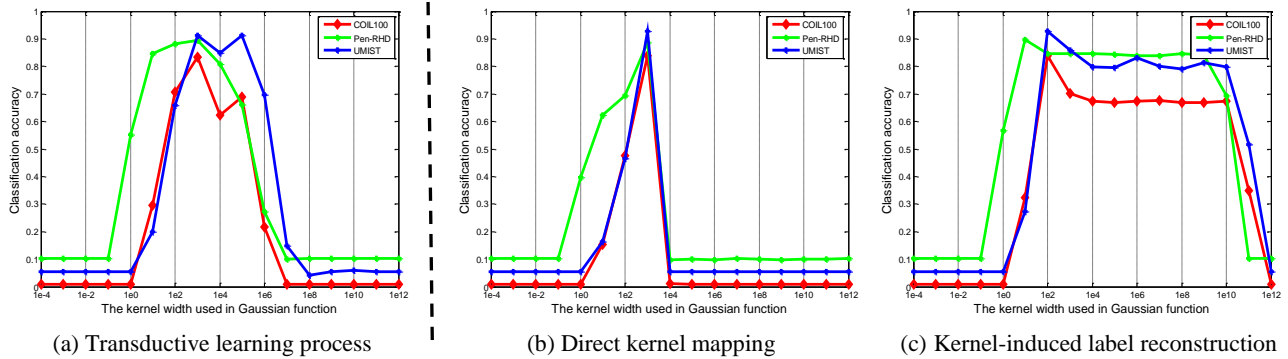

(a) Transductive learning process    (b) Direct kernel mapping    (c) Kernel-induced label reconstruction

**Fig. 10:** Sensitivity analysis of different selections of kernel width $\sigma$ on the three real-world databases.

**(2) Investigation of the Gaussian kernel width $\sigma$.** In this experiment, we will present two groups of analysis results, i.e., exploring the effects of kernel width $\sigma$ from the candidate set $\{10^{-4}, 10^{-3}, \ldots, 10^{11}, 10^{12}\}$ on the transductive and inductive processes of our Kernel-LP. For our inductive process, we also present two results based on the direct kernel mapping and the kernel-induced label reconstruction.

The selection results of kernel width $\sigma$ on the transductive learning process are shown in Fig.10 (a), from which we can observe that the best results are all obtained at $\sigma=10^3$ for the tested three kinds of databases. Thus, we simply set $\sigma=10^3$ for our Kernel-LP for semi-supervised classification.

The selection results of kernel width $\sigma$ on the direct kernel mapping and kernel-induced label reconstruction process are illustrated in Fig.10 (b-c) respectively. From the results on the direct kernel mapping, we can find that the best records are also obtained at $\sigma=10^3$ for the tested three kinds of databases. While for the analysis on kernel-induced label reconstruction process, one can easily see that promising results are usually achieved at $\sigma=10^2$ over the tested databases. Thus, we can set the width $\sigma$ to $10^3$ and $10^2$ for our inductive I-Kernel-map and I-Kernel-recons to include the outside new data.

## VI. CONCLUDING REMARKS

We have proposed a novel kernel-induced label propagation framework termed Kernel-LP by mapping for semi-supervised classification. The core idea of our Kernel-LP is to change the scenario of label propagation from commonly-used Euclidean distance to kernel space so that more informative patterns and relations of samples can be accurately discovered for learning useful knowledge based on the mapping assumption of kernel trick. Thus, the similarity can be encoded more accurately for enhancing subsequent representation. For kernel-induced label propagation, the seamless integration of adaptive graph weight construction and kernelized label propagation can also ensure the weights to be joint-optimal for data representation and classification in kernel space. The positive effects of negative label information are also used to boost the results. To enable Kernel-LP to process new data efficiently, two novel out-of-sample methods by direct kernel mapping and kernel-induced label reconstruction are also presented. The proposed new data inclusion methods only depends on the kernel matrix between training set and testing set, which is simple and efficient.

Extensive results on image recognition and segmentation have demonstrated the effectiveness of our kernelized methods for the transductive and inductive classification. Extending the kernelized scheme of our algorithm to other related methods will be investigated in future. Besides, extending our methods to the other application areas, such as image retrieval and text categorization, is also worth investigating in future.

## ACKNOWLEDGMENT

This work is partially supported by National Natural Science Foundation of China (61672365, 61502238, 61622305, 61432019, 61772171, 61601112), Major Program of Natural Science Foundation of Jiangsu Higher Education Institutions of China (15KJA520002), "Qing-Lan Project" of Jiangsu Province, Natural Science Foundation of Jiangsu Province of China (BK20160040), and High-Level Talent of "Six Talent Peak" Project of Jiangsu Province of China (XYDXX-055).

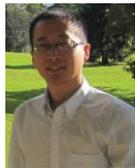
**Zhao Zhang** (M'13-SM'17-) received the Ph.D. degree from the Department of Electronic Engineering (EE), City University of Hong Kong, in 2013. He is now an Associate Professor at School of Computer Science and Technology, Soochow University, Suzhou, P. R. China. Dr. Zhang was a Visiting Research Engineer at the National University of Singapore, from Feb to May 2012. He then visited the National Laboratory of Pattern Recognition (NLPR) at Chinese Academy of Sciences (CAS), from Sep to Dec 2012. His current research interests include data mining & statistical machine learning, pattern recognition & image analysis. He has authored/co-authored about 70 technical papers published at prestigious international journals and conferences, including IEEE TNNLS (3), IEEE TIP (3), IEEE TKDE (3), IEEE TCYB, IEEE TSP, IEEE TII (2), ACM TIST, Pattern Recognition (5), Neural Networks (5), ICDM, ACM ICMR, ICIP and ICPR, etc. Specifically, he has got 12 first-authored regular papers published by IEEE/ACM Transactions journals. Dr. Zhang is now serving as an Associate Editor (AE) of Neurocomputing, IET Image Processing, and Neural Processing Letters (NPL); also has served as an editorial board member for Neural Computing and Applications (NCA). Besides, he served as a Senior Program Committee (SPC) member of PAKDD 2018/2017, an Area Chair (AC) of BMVC 2016/2015, a Program Committee (PC) member for several popular international conferences (e.g., AAAI 2018, SDM 2017/2016/2015, PRICAI 2018, PAKDD 2016, IJCNN 2017/2015, CAIP 2015, etc), and also often got invited as a journal reviewer for IEEE TNNLS、IEEE TIP、IEEE TKDE、IEEE TSP、IEEE TCYB、IEEE TMM, IEEE TII、Pattern Recognition, etc. He is now a Senior Member of the IEEE.

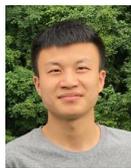
**Lei Jia** is currently working toward the research degree at the School of Computer Science and Technology, Soochow University, China. His current research interests include machine learning, pattern recognition, data mining and their real applications. More specifically, he is interested in designing effective semi-supervised learning algorithms for classification. He has published several papers in Neural Networks, and conference proceedings of ICDM, ICONIP and PCM.

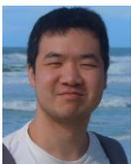
**Mingbo Zhao** (M'13- ) received the Ph.D. degree from the Department of Electronic Engineering at City University of Hong Kong, Kowloon, Hong Kong SAR, in 2013. He is now a Senior Researcher at the City University of Hong Kong. His current research interests mainly include pattern recognition, machine learning and data mining. He has authored/co-authored more than 50 papers published at prestigious international journals and conferences, e.g., IEEE TIP, IEEE TII, IEEE TKDE, ACM TIST, Pattern Recognition, Computer Vision and Image Understanding (CVIU), Information Sciences, Neural Networks, etc.

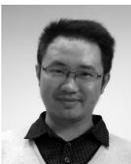
**Guangcan Liu** (M'11- ) received the bachelor's degree in mathematics and the Ph.D. degree in computer science and engineering from Shanghai Jiao Tong University, Shanghai, China, in 2004 and 2010, respectively. He was a Post-Doctoral Researcher with the National University of Singapore, Singapore, from 2011 to 2012, the University of Illinois at Urbana-Champaign, Champaign, IL, USA, from 2012 to 2013, Cornell University, Ithaca, NY, USA, from 2013 to 2014, and Rutgers University, Piscataway, NJ, USA, in 2014. Since 2014, he has been a Professor with the School of Information and Control, Nanjing University of Information Science and Technology, Nanjing, China. His research interests touch on the areas of pattern recognition and signal processing. He obtained the National Excellent Youth Fund in 2016 and was designated as the global Highly Cited Researchers in 2017.

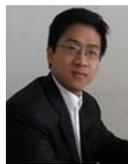
**Meng Wang** is a professor in the Hefei University of Technology, China. He received the B.E. degree and Ph.D. degree in the Special Class for the Gifted Young and signal and information processing from the University of Science and Technology of China (USTC), Hefei, China, respectively. He previously worked as an associate researcher at Microsoft Research Asia, and then a core member in a startup in Bay area. After that, he worked as a senior research fellow in National University of Singapore. His current research interests include multimedia content analysis, search, mining, recommendation, and large-scale computing. He has authored more than 200 book chapters, journal and conference papers in these areas, including TMM, TNNLS, TCSVT, TIP, TOMCCAP, ACM MM, WWW, SIGIR, ICDM, etc. He received the paper awards from ACM MM 2009 (Best Paper Award), ACM MM 2010 (Best Paper Award), MMM 2010 (Best Paper Award), ICIMCS 2012 (Best Paper Award), ACM MM 2012 (Best Demo Award), ICDM 2014 (Best Student Paper Award), PCM 2015 (Best Paper Award), SIGIR 2015 (Best Paper Honorable Mention), IEEE TMM 2015 (Best Paper Honorable Mention), and IEEE TMM 2016 (Best Paper Honorable Mention). He is the recipient of ACM SIGMM Rising Star Award 2014. He is the recipient of ACM SIGMM Rising Star Award 2014. He is/has been an Associate Editor of IEEE Trans. on Knowledge and Data Engineering (TKDE), IEEE Trans. on Neural Networks and Learning Systems (TNNLS) and IEEE Trans. on Circuits and Systems for Video Technology (TCSVT). He is a member of IEEE and ACM.

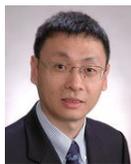
**Shuicheng Yan** (F'16- ) is now a (Dean's Chair) Associate Professor in the Department of Electrical and Computer Engineering at National University of Singapore, and the founding lead of the Learning and Vision Research Group (http://www.lv-nus.org). Dr. Yan's research areas include computer vision, multimedia and machine learning, and he has authored/co-authored over 370 technical papers over a wide range of research topics, with Google Scholar citation >12,000 times and H-index-47. He is an associate editor (AE) of Journal of Computer Vision and Image Understanding, IEEE Trans. Knowledge and Data Engineering (TKDE), IEEE Trans. Circuits and Systems for Video Technology (IEEE TCSVT) and ACM Trans. Intelligent Systems and Technology (ACM TIST), and has been serving as the guest editor of the special issues for TMM and CVIU. He received the Best Paper Awards from ACM MM13 (Best Paper and Best Student Paper), ACM MM'12 (demo), PCM'11, ACM MM'10, ICME'10 and ICIMCS'09, the winner prizes of the classification task in PASCAL VOC 2010-2012, the winner prize of the segmentation task in PASCAL VOC 2012, the honorable mention prize of the detection task in PASCAL VOC'10, 2010 TCSVT Best Associate Editor (BAE) Award, 2010 Young Faculty Research Award, 2011 Singapore Young Scientist Award, 2012 NUS Young Researcher Award, and the co-author of the best student paper awards of PREMIA'09, PREMIA'11 and PREMIA'12. He shall or has been General/Program Co-chair of PCM13, and ACM MM15, ICMR17 and ACM MM17.